\documentclass[]{gtech}

\usepackage{amssymb}
\usepackage{multirow}
\usepackage{bigdelim}
\usepackage{todonotes}
\usepackage{longtable}
\usepackage{tabularray}
\usepackage{wrapfig}
\usepackage[most]{tcolorbox} 
\usepackage{xcolor}
\usepackage{url}
\usepackage{algorithm}
\usepackage{algpseudocode} 
\usepackage[toc,page]{appendix}
\usepackage{multirow}
\usepackage[normalem]{ulem}
\usepackage{adjustbox}
\usepackage{booktabs}  
\usepackage{natbib}
\usepackage{colortbl}  
\usepackage{subcaption}
\usepackage{pifont}
\usepackage{amsmath}
\usepackage{amsfonts}       
\usepackage{multirow}
\usepackage{mathrsfs}
\usepackage{tabularx}
\usepackage[utf8]{inputenc} 
\usepackage[T1]{fontenc}    
\usepackage{hyperref}       
\usepackage{url}            
\usepackage{booktabs}       
\usepackage{nicefrac}       
\usepackage{microtype}      
\usepackage{xcolor}         
\definecolor{ourreward}{HTML}{32bbee}
\definecolor{sparsereward}{HTML}{ed9a81}
\usepackage{graphicx}
\usepackage{wrapfig}
\usepackage{cleveref}

\usepackage{bbding}

\useunder{\uline}{\ul}{}
\newcommand*{\affaddr}[1]{#1}

\title{%
  \texorpdfstring{%
    \raisebox{-0.23\height}{\includegraphics[scale=0.03]{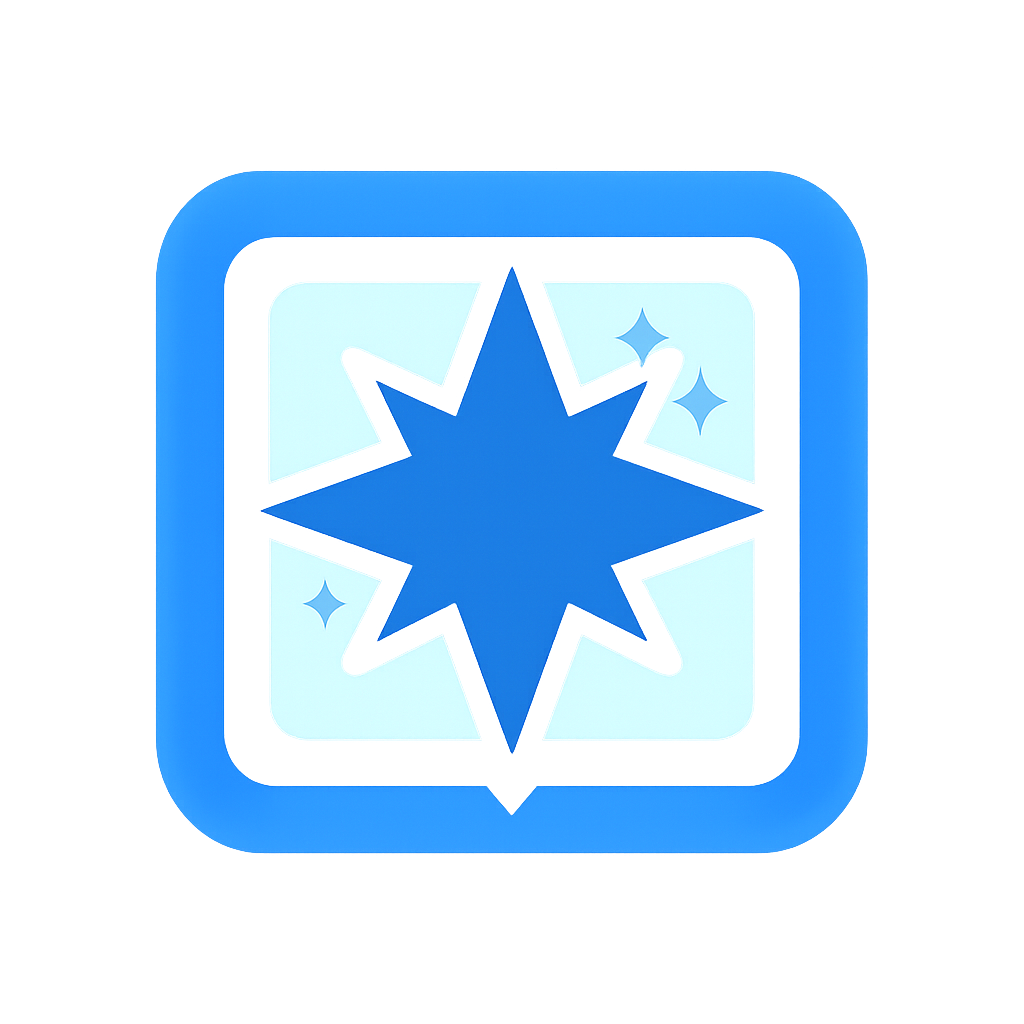}}%
    ~UI-Venus Technical Report: \\ Building High-performance UI Agents with RFT%
  }{UI-Venus Technical Report: \\ Building High-performance UI Agents with RFT}%
}

\author{%
\small
Zhangxuan~Gu$^*$, Zhengwen~Zeng$^*$, Zhenyu~Xu$^*$, Xingran~Zhou$^*$,  Shuheng~Shen$^{*,{\dag}}$, Yunfei~Liu$^*$, Beitong~Zhou$^*$, Changhua~Meng, Tianyu~Xia, Weizhi~Chen, Yue~Wen, Jingya~Dou, Fei~Tang, Jinzhen~Lin, Yulin~Liu, Zhenlin~Guo, Yichen~Gong, Heng~Jia, Changlong~Gao, Yuan~Guo, Yong~Deng,  Zhenyu~Guo, Liang~Chen, Weiqiang~Wang \\
\affaddr{Ant Group}
}

\abstract{
We present UI-Venus, a native UI agent that takes only screenshots as input based on a multimodal large language model. UI-Venus achieves SOTA performance on both UI grounding and navigation tasks using only several hundred thousand high-quality training samples through reinforcement finetune (RFT) based on Qwen2.5-VL. 
Specifically, the 7B and 72B variants of UI-Venus obtain \textbf{94.1\%} / \textbf{50.8\%} and \textbf{95.3\%} / \textbf{61.9\%} on the standard grounding benchmarks, \emph{i.e.}, Screenspot-V2 / Pro, surpassing the previous SOTA baselines including open-source GTA1 and closed-source UI-TARS-1.5.
To show UI-Venus's summary and planing ability, we also evaluate it on the AndroidWorld, an online UI navigation arena, on which our 7B and 72B variants achieve \textbf{49.1\%} and \textbf{65.9\%} success rate, also beating existing models.
To achieve this, we introduce carefully designed reward functions for both UI grounding and navigation tasks and corresponding efficient data cleaning strategies.
To further boost navigation performance, we propose Self-Evolving Trajectory History Alignment \& Sparse Action Enhancement that refine historical reasoning traces and balances the distribution of sparse but critical actions, leading to more coherent planning and better generalization in complex UI tasks. 
Our contributions include the publish of SOTA open-source UI agents, comprehensive data cleaning protocols and a novel self-evolving framework for improving navigation performance, which encourage further research and development in the community.
}

\gtechdata[Code\&Model]{\url{https://github.com/inclusionAI/UI-Venus}}

\begin{document}
    \maketitle

\renewcommand{\thefootnote}{}

\footnote{*Equal contribution. $^{\dag}$Corresponding author: Shuheng~Shen(shuheng.ssh@antgroup.com). }

\begin{figure}[htbp]
    \centering
    \begin{subfigure}{1.0\textwidth}
        \centering
        \vspace{-8mm}
        \includegraphics[width=\textwidth]{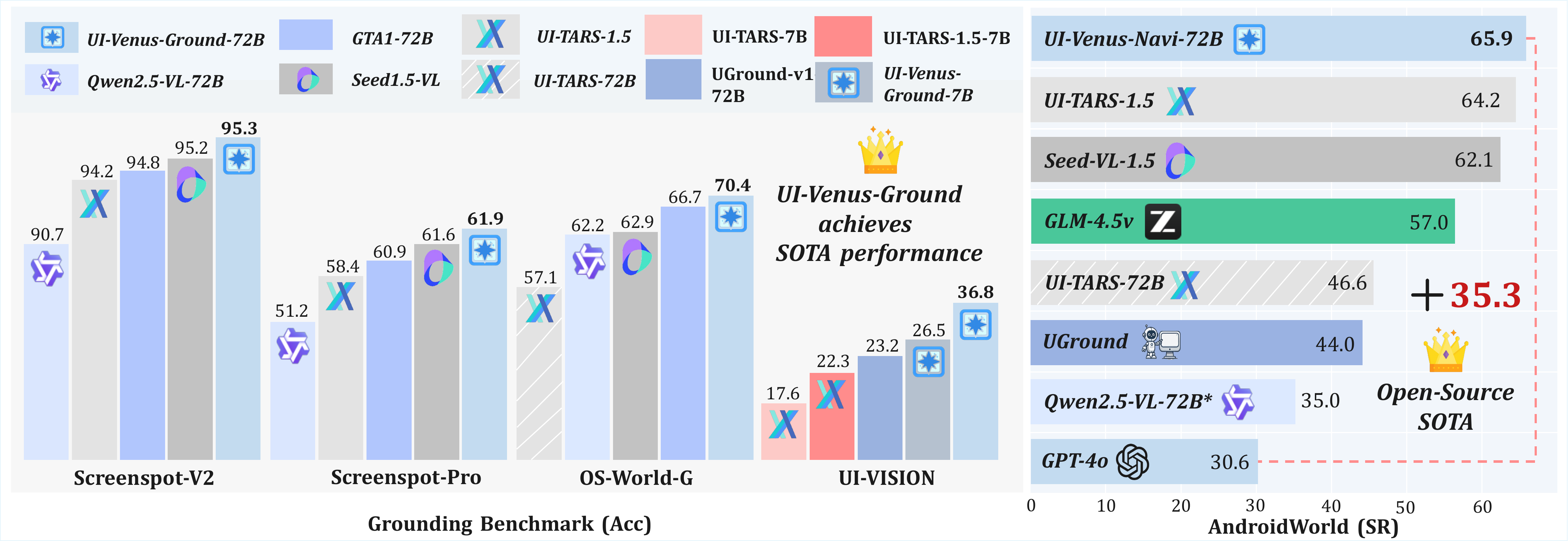}
    \end{subfigure}
    \caption{UI-Venus achieves SOTA performance across multiple UI grounding and navigation benchmarks.}
    \label{fig:performance_gd}
\end{figure}



\section{Introduction}
Recent studies of multimodal large language models (MLLMs)~\cite{bai2023qwenvlversatilevisionlanguagemodel,anthropic2024cuda,wang2024qwen2vlenhancingvisionlanguagemodels,bai2025qwen25vltechnicalreport,zhu2025internvl3exploringadvancedtraining,glm-4.5v} have contributed significantly to the advancements and developments of UI agents~\cite{hu2024agents,gao2024generalist,wang2024gui,nguyen2024gui,zhang2024large}. In particular, many early approaches, \emph{e.g.}, CogAgent~\cite{hong2024cogagentvisuallanguagemodel} and UI-TARS~\cite{qin2025uitarspioneeringautomatedgui}, directly leverage extensive open-source and private datasets through pretraining and supervised fine-tuning (SFT) to achieve commendable performance in UI agent tasks. Specifically, these pretrained and SFT approaches treat the complete UI task traces, state observations, and current actions as plain texts for token-level supervised learning. Although SFT is effective in many generation tasks with strong instruction-following ability, sometimes it is not appropriate in discriminative tasks like UI agent. 
For example, in UI grounding tasks, a predicted point is considered correct as long as it falls within the ground-truth bounding box. In contrast, SFT assigns widely-used cross-entropy loss penalty to any predicted points within and without the box except the center point. Moreover, the data collection and cleaning workflow during pretraining is time consuming and requires a lot of human work.

Inspired by the emergence of DeepSeek-R1~\cite{deepseekai2025deepseekr1incentivizingreasoningcapability} and its innovative Group Relative Policy Optimization (GRPO) algorithm~\cite{shao2024deepseekmathpushinglimitsmathematical}, recent researchers focus on reinforcement finetune (RFT) paradigms for discriminative tasks such as math and code. RFT generally needs fewer training data, yet has better generalization abilities compared to SFT, as exemplified by~\cite{chen2025sftrlearlyinvestigation}. 
As a result, many approaches similar to UI-R1~\cite{lu2025uir1enhancingefficientaction} have also achieved satisfactory results on UI grounding benchmarks by adapting and modifying the reward functions of VLM-R1~\cite{shen2025vlmr1stablegeneralizabler1style}. For example, GUI-G1~\cite{zhou2025guig1understandingr1zeroliketraining}, GUI-G$^2$~\cite{guig2}, Phi-Ground~\cite{phiground} and GTA1~\cite{yang2025gta1guitesttimescaling} mainly focus on UI grounding benchmarks (ScreenSpotv2~\cite{wu2024osatlasfoundationactionmodel} and ScreenSpotpro~\cite{li2024screenspot-pro}) with different data collection and ratios. Some tricks like think / no-think and input message construction are also discussed in their reports. On the other hand, GUI-R1~\cite{luo2025guir1generalistr1style}, InfiGUI-R1~\cite{liu2025infiguir1advancingmultimodalgui} and UI-R1~\cite{lu2025uir1enhancingefficientaction} extend GRPO to UI offline navigation benchmarks (Andorid Control~\cite{androidcontrol} and GUI-Odyssey~\cite{guiodyssey}) by adapting action type, format and the point rewards with predefined action spaces.



Despite these advancements, current R1-like UI agents exhibit three critical limitations. First, although some attempts are made by UI-R1\cite{lu2025uir1enhancingefficientaction} and InfiGUI-R1~\cite{liu2025infiguir1advancingmultimodalgui} in offline UI navigation tasks, their summary, memory and planning ability are still not enough for online tasks like AndroidWorld~\cite{androidworld} when given a user goal with complex and changeful interactive environment. 
Secondly, one important factor for training UI agents is the data quality. According to our observations, approximately a half of open-source UI data contain noise, while only a few methods have considered the data cleaning and selection strategies.
Thirdly, existing implementations mainly focus on small MLLMs (\emph{e.g.}, 3B/7B parameters), neglecting the potential of large-scale models (\emph{e.g.}, 72B) in RFT training. Although GTA1 has developed 72B grounding model, no further attempts are made on the end-to-end UI navigation tasks without addtional planner like GPT4o~\cite{openai2024gpt4o}. This constraint results in performance gaps compared to state-of-the-art large-scale models like UI-TARS-1.5~\cite{ui-tars-15-seed} and SeedVL-1.5~\cite{seed2025seed1_5vl} on UI agent benchmarks. Moreover, the evaluation for UI agents on many benchmarks suffers from severe challenges due to different input prompts as well as unreleased hyper-parameters and dataset settings. For example, some results from the official papers can not be reproduced according to \cite{yang2025gta1guitesttimescaling,agentcpm} and some github issues.

To address the first challenge mentioned above, we design a Self-Evolving Trajectory History Alignment \& Sparse Action Enhancement framework. This method addresses two critical limitations in existing UI navigation agents: (1) misaligned historical reasoning traces and (2) insufficient learning of rare but pivotal actions. For the first issue, we iteratively refine the thought–action histories between training epochs, aligning the form and level of detail of the historical reasoning with the agent’s evolving decision-making patterns. This produces a more coherent and informative context for predicting subsequent steps, which in turn improves planning accuracy. For the second, we selectively re-sample trajectories containing sparse actions, constructing multiple historical context variants that lead to the same low-frequency operation. This balanced sampling increases the model’s exposure to infrequent yet crucial skills, enhancing its ability to generalize in complex and dynamic UI task scenarios.

To acquire high-quality UI data, we implement a three-stage processing pipeline to tackle the second problem: (1) \textbf{Data Filtering} includes unifying scroll directions, filtering out trajectories with inconsistencies, and resampling trajectories based on their categories. (2) \textbf{Trace Reconstruction} refers to modifying the information-retrieval traces with specific answers inserted. (3) \textbf{Iteratively Trace Generation:} we develop a data generation framework by using UI-Venus-Navi to predict and record trajectories, with comprehensive quality filtering strategies to select high-quality traces for iterative training. During the training, we use about 107k/350k high-quality training samples filtered by ourselves for UI grounding/navigation task, respectively. More details will be introduced in the next section.

In this report, we develop and publicly release the UI-Venus series, comprising UI-Venus-Ground-7B/72B and UI-Venus-Navi-7B/72B, all trained with GRPO on the Qwen2.5-VL model~\cite{bai2025qwen25vltechnicalreport}. We also open-source the evaluation codes of grounding as well as the prompts and post-processing scripts of navigation to enhance accessibility and ease of use for researchers.
Experimental results exhibit that UI-Venus outperforms all existing UI agents, showing strong performance with about 350k self-constructed high quality dataset. Our model achieves new SOTA performance on five UI grounding benchmarks including ScreenSpot-V2~\cite{wu2024osatlasfoundationactionmodel}, ScreenSpot-Pro~\cite{li2024screenspot-pro}, OSWorld-G~\cite{xie2025scalingcomputerusegroundinguser}, UI-Vision~\cite{nayak2025ui} and CA-GUI~\cite{agentcpm}.
The 7B and 72B variants of UI-Venus obtain \textbf{94.1\%} / \textbf{50.8\%} and \textbf{95.3\%} / \textbf{61.9\%} on the Screenspot-V2 / Pro, surpassing previous SOTA baselines including open-source GTA1 and closed-source UI-TARS-1.5. For UI navigation, we evaluate UI-Venus on both offline and online UI navigation benchmarks. Our model achieves SOTA performance on AndroidWorld~\cite{androidworld}, where 7B and 72B variants achieve \textbf{49.1\%} and \textbf{65.9\%} success rate, while obtaining comparable results on AndroidControl~\cite{androidcontrol} and GUI-Odyssey~\cite{guiodyssey} benchmarks.

There are two motivations for publishing the grounding and navigation models separately. 
(1) \textbf{A More Efficient Grounding Model:} According to our experiments, explicitly prompting the model to output its thinking process significantly enhances its observation and planning capabilities in UI navigation tasks. However, UI-Venus achieves approximate performance on UI grounding tasks with and without thinking. As no-think mode for grounding outputs only several location tokens, it's more efficient in inference for real-world applications.
(2) \textbf{Reward Conflicts:} If we attempt to train an agent using both grounding and navigation tasks, a feasible approach is to convert grounding data into single-step click actions in the navigation task. However, using these two reward mechanisms simultaneously can lead to unstable training, ultimately degrading performance on both tasks.


We summarize our main contributions as follows:
\begin{itemize}
    \item To mitigate historical reasoning misalignment and augment the learning of rare but pivotal actions, we design a Self-Evolving Trajectory History Alignment \& Sparse Action Enhancement framework, which in turn boosts navigation performance in complex UI scenarios.
    \item We conduct a comprehensive study of the quality of UI data, and propose a data cleaning and selection strategy to improve training data quality for both grounding and navigation.
    \item We develop and open-source UI-Venus, a SOTA UI agent for both grounding and navigation tasks with carefully designed reward functions, whose 7B and 72B variants obtaining \textbf{94.1\%} / \textbf{50.8\%} and \textbf{95.3\%} / \textbf{61.9\%} on Screenspot-V2 / Pro,  and \textbf{49.1\%} and \textbf{65.9\%} on AndroidWorld, demonstrating the effectiveness of model scaling up for GRPO in UI-agent tasks.
\end{itemize}

\section{Related Works}

\subsection{UI Grounding} 
\vspace{2mm}
UI Grounding focuses on localizing and identifying UI elements based on natural language instructions, serving as the foundation of automated UI interaction. Traditional approaches have relied on Supervised Fine-Tuning to train models on labeled UI datasets~\cite{cheng2024seeclickharnessingguigrounding,lin2024showuivisionlanguageactionmodelgui,xu2024aguvis,wu2024osatlasfoundationactionmodel,gu2023mobile,gou2024navigatingdigitalworldhumans,wang2024eantlargescaledatasetefficient}. However, these methods face two primary limitations: (1) poor generalization in out-of-distribution scenarios where UI layouts or visual styles differ from training data, and (2) high costs associated with acquiring large-scale annotated datasets for diverse UI environments. Recent advances have shifted toward reinforcement learning-based fine-tuning inspired by the DeepSeek-R1 paradigm. Early works like UI-R1~\cite{lu2025uir1enhancingefficientaction} and GUI-R1~\cite{luo2025guir1generalistr1style} introduced binary hit-or-miss rewards for task completion, while InfiGUI-R1~\cite{liu2025infiguir1advancingmultimodalgui} proposed two-stage training combining offline pretraining with online RL. GUI-G1~\cite{zhou2025guig1understandingr1zeroliketraining} further refined the reward design with box-size-based rewards for improved spatial precision. The latest methods, including SE-GUI~\cite{yuan2025enhancingvisualgroundinggui}, LPO~\cite{tang2025lpoaccurateguiagent}, and GUI-G$^2$~\cite{guig2}, employ continuous reward mechanisms that provide fine-grained feedback throughout the grounding process. 

\subsection{UI Agents}
\vspace{2mm}
\noindent \textbf{UI Agent Framework.} 
The UI agent framework leverages collaborative agent systems to handle complex GUI automation tasks through task decomposition and specialization. Mobile-Agent~\cite{wang2024mobile,wang2024mobileagentv2mobiledeviceoperation,wang2025mobileagenteselfevolvingmobileassistant} addresses mobile device automation by introducing a planning-decision-reflection architecture, where specialized agents handle task progress tracking, action execution, and operation verification respectively, while maintaining a memory unit for context preservation across interactions. Cradle~\cite{tan2024cradle} presents a modular framework with six core components—information gathering, self-reflection, task inference, skill curation, action planning, and memory—enabling agents to tackle both video game control and software manipulation through adaptive learning and skill reuse. Agent-S~\cite{agashe2024agent} implements role-specific agents that specialize in distinct UI interaction patterns, improving task execution through modular decision-making. DroidRun~\cite{droidrun} focuses on Android automation by leveraging Accessibility Services to access structured UI hierarchies rather than pixel-based approaches, enabling reliable interaction with native mobile applications
. These frameworks illustrate that agent specialization, in conjunction with structured communication protocols, can effectively address diverse UI scenarios. However, this capability comes at the cost of increased system complexity and computational overhead required to coordinate multiple agents.

\noindent \textbf{Native UI Agent.}  
Native UI agents~\cite{hong2024cogagentvisuallanguagemodel,qin2025uitarspioneeringautomatedgui,agentcpm,phiground,feng2025efficientonlinetuningvlm,bai2024digirltraininginthewilddevicecontrol,humphreys2022data} represent a paradigm shift toward unified and end-to-end systems that directly learn to interact with graphical interfaces without requiring multiple specialized components. CogAgent~\cite{hong2024cogagentvisuallanguagemodel} and UI-TARS~\cite{qin2025uitarspioneeringautomatedgui} pioneered this approach through large-scale training on diverse UI interaction data, enabling the model to develop a comprehensive understanding of GUI patterns across different platforms and applications. By training on millions of UI screenshots paired with action sequences, UI-TARS demonstrated that a single model could effectively handle various tasks. Following this success, the native UI agent introduced architectural improvements that enhance the model's ability to process high-resolution screenshots while maintaining efficient action prediction, particularly excelling in scenarios with dynamic UI updates and real-time feedback requirements. AgentCPM-GUI~\cite{agentcpm} took a different approach by focusing on mobile-specific optimizations, incorporating touch gesture understanding and mobile UI design patterns into its training process, resulting in superior performance on mobile platforms with reduced latency. Meanwhile, Phi-Ground~\cite{phiground} advanced the field by integrating sophisticated vision-language alignment techniques, enabling more robust grounding of natural language instructions to visual UI elements through cross-modal attention mechanisms. These native agents benefit from their streamlined architecture and ability to learn complex UI interaction patterns directly from data, though they face challenges in data efficiency and often require substantial computational resources for training on diverse UI environments.

\section{Methodology}

\begin{figure}[t]
\centering
\includegraphics[width=1.0\linewidth]{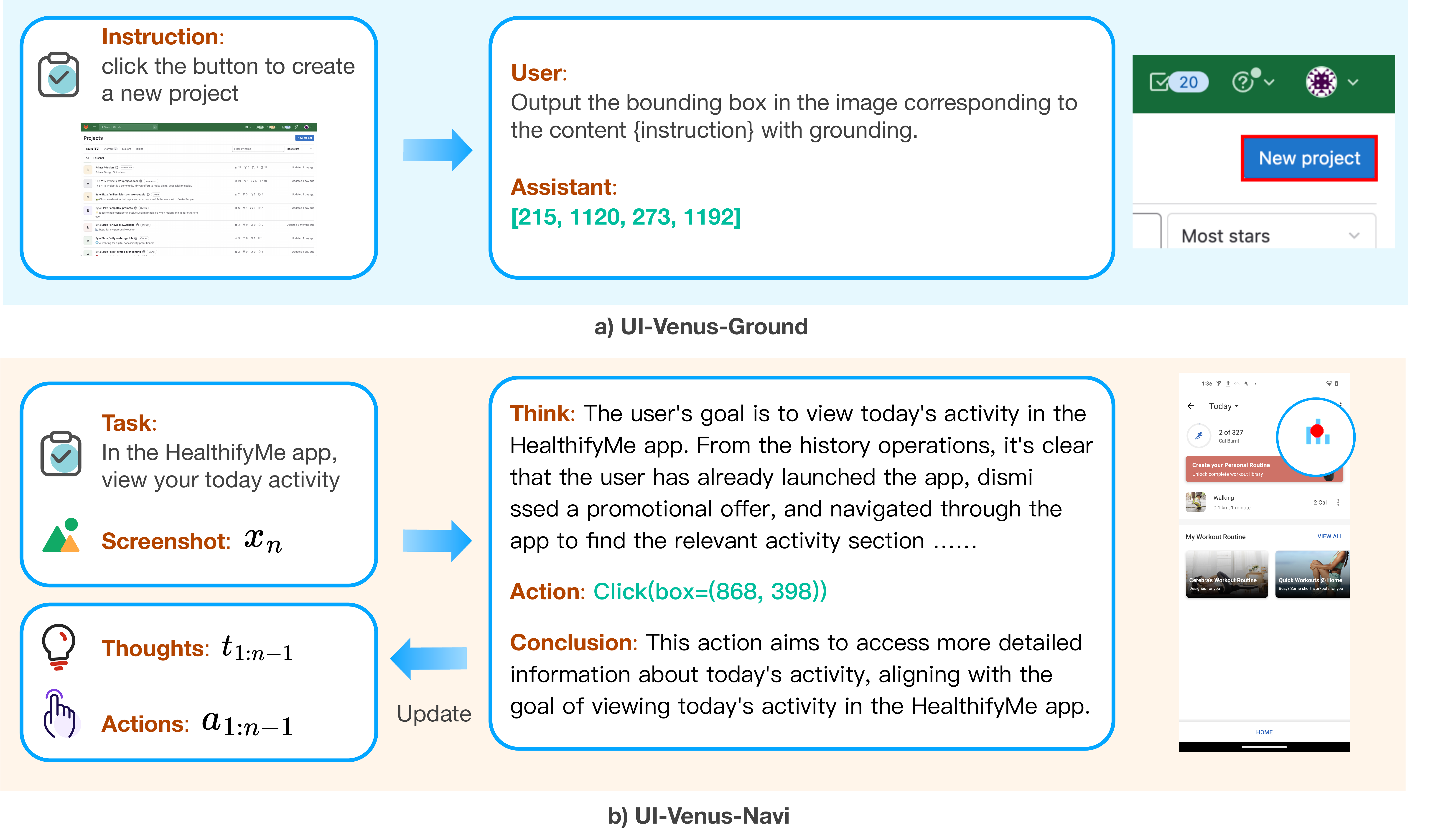}
\caption{Executions on typical grounding and navigation tasks of UI-Venus. \textbf{a)} The instruction and the screenshot are needed for UI-Venus-Ground to output the corresponding coordinates; \textbf{b)} For navigation tasks, historical context (thought-action pairs) are essential for UI-Venus-Navi, which will generate the thinking content and model action. The historical context will be updated after each step finished.}
\label{fig:ui_venus_framework}
\end{figure}

Many prior works have successfully verified that Reinforcement Fine-Tune (RFT) based on Group Relative Policy Optimization (GRPO) algorithm is suitable for UI grounding, where the MLLM answers a response that includes the predicted boxes given corresponding objective descriptions. In this work, we further extend GRPO to UI navigation task and prove its effectiveness in UI Agents training compared to merely SFT. We first present the preliminaries of GRPO, followed by a comprehensive discussion of the data pipelines, reward design, and learning framework in UI grounding and UI navigation respectively.

\subsection{Preliminaries}
GRPO enhances training stability by estimating baselines through relative rewards within groups rather than using a separate critic model. For each training question $q \in Q$, GRPO samples $G$ rollouts $\{o_1, o_2, \ldots, o_G\}$ from the current policy $\pi_\theta$ and computes their corresponding rewards $\{r_1, r_2, \ldots, r_G\}$. The core idea lies in normalizing rewards within each group to obtain advantages: $\hat{A}_i = \frac{r_i - \text{mean}(\{r_1, r_2, \ldots, r_G\})}{\text{std}(\{r_1, r_2, \ldots, r_G\})}$. The policy is then optimized by maximizing the objective:
{
\small
\begin{align}
\mathcal{J}_{\mathrm{GRPO}}(\pi_{\theta}) &= \mathbb{E}_{q\sim \mathcal{Q}, {\{o_i\}_{i=1}^G\sim \pi_{\theta_{\mathrm{old}}}(\cdot|q)}}\notag\\
\frac{1}{G}\sum_{i=1}^{G}\frac{1}{|o_i|}&\sum_{t=1}^{|o_i|} \left\{\min \left [\frac{\pi_{\theta}(o_{i,t}|q,o_{i,<t})}{\pi_{\theta_{\scriptstyle \mathrm{old}}}(o_{i,t}|q,o_{i,<t})}\hat{A}_{i}, \mathrm{clip}\Big (\frac{\pi_{\theta}(o_{i,t}|q,o_{i,<t})}{\pi_{\theta_{\scriptstyle \mathrm{old}}}(o_{i,t}|q,o_{i,<t})}, 1\!-\!\epsilon, 1\!+\!\epsilon\Big ) \hat{A}_{i} -\beta \mathrm{D}_{\mathrm{KL}}\left[\pi_\theta\parallel\pi_{\mathrm{ref}}\right]\right ]\right\}, \label{eq:grpo}\\
& \qquad\qquad\qquad\mathrm{where}\quad \hat{A}_i = \frac{r_i - \mathrm{mean}(\{r_1, r_2, \cdots r_G\})}{\mathrm{std}(\{r_1, r_2, \cdots r_G\})},\notag
\end{align}
}where the clipping mechanism prevents excessive policy updates, $\epsilon$ controls the clipping range, and the KL divergence term with coefficient $\beta$ constrains the policy from diverging excessively from the reference model $\pi_{\text{ref}}$.

\subsection{UI Grounding}

\subsubsection{Data Collection}
Preliminarily, we collect UI grounding instances from existing public datasets, including Widget Captioning~\cite{cheng2024seeclickharnessingguigrounding}, UI RefExp~\cite{bai2021uibertlearninggenericmultimodal}, SeeClick~\cite{cheng2024seeclickharnessingguigrounding}, ShowUI~\cite{lin2024showuivisionlanguageactionmodelgui}, and OmniAct~\cite{ominiact}. They are extracted from different platforms such as mobile, desktop and web to ensure the diversity of training data.
As shown in Table~\ref{tab:grounding_data}, we collect about 627k open-source grounding samples and use about 107k for training after carefully data cleaning.

\begin{table*}[h]\small
\centering
\renewcommand{\arraystretch}{1.1}
\setlength{\tabcolsep}{6pt}
\begin{tabular}{c|ccccc|c}
\toprule
\textbf{} & \textbf{Widget Captioning} & \textbf{UI RefExp} & \textbf{SeeClick-Web} & \textbf{ShowUI} & \textbf{OmniAct} & \textbf{Sampled training set} \\
\midrule
\textbf{\# Samples} & 34k & 17k & 325k & 110k & 141k  & \textbf{107k} \\
\bottomrule
\end{tabular}
\vspace{0.4em}
\caption{GUI grounding data composition. Our training dataset comprises 107k samples from five complementary sources.}
\label{tab:grounding_data}
\end{table*}

\subsubsection{Data Cleaning}

However, according to our observation, approximately 40\% of the current open-source grounding data contain significant noise issues, including prompt ambiguity and box shifts. A possible cause is that many web and mobile datasets utilize HTML and A11Y (Accessibility) source code, respectively. Although these source codes contain the bounding boxes and their corresponding descriptive instructions required for grounding tasks, some boxes may have mismatched offsets after page rendering. In severe cases, all boxes in one entire image may shift simultaneously in one certain direction. Additionally, due to nested elements in UI components from source code, bounding boxes and instructions may fail to maintain a one-to-one correspondence, with many-to-one and one-to-many mappings potentially affecting model training.

Since RFT training requires high-quality clean data, we employ manual inspection to ensure the correctness of box-instruction pairs. Specifically, given the original open source datasets (\emph{e.g.}, Seeclick), which are large in scale but contain many redundant or simple samples, we first perform downsampling to create a subset by removing the repeated prompts. Then we manually filter out ambiguous prompts, relocate offset boxes, and rephrase unmeaningful instructions. After these steps, we obtain about 107k high-quality training samples, which is shown in Table \ref{tab:grounding_data}.

Failed attempts for automatic grounding data cleaning include: (1) Using open-source models to perform the above filtering and rewriting operations; (2) RFT with hard samples selected by rejection sampling strategies.
Although these approaches failed to work, they provide some insight for the UI-agent community.

\subsubsection{Reward Function}

In the UI grounding task, we only use two reward functions, \emph{i.e.}, point-in-box and format reward. Although many approaches~\cite{lu2025uir1enhancingefficientaction,guig2,luo2025guir1generalistr1style} use intersection-over-union (IoU) or other smooth rewards, the results are similar according to our experiments. Specifically, the reward function is composed of two components formally:

\noindent\textbf{Format Reward. }We first check whether the predicted answer string conforms to a predefined syntax. Valid answers receive a base reward to ensure the model to generate executable and parsable instructions.

\noindent\textbf{Point-in-box Reward. }Given a screenshot and the instruction, the model must predict a bounding box that localizes the element, where $(x_c, y_c)$ denote the box center. Assume that the ground truth is annotated as $[x_1, y_1, x_2, y_2]$, then

\begin{equation}
R_{\text{poin-in-box}} =
\begin{cases}
1 & \text{if }  x_1\le x_c \le x_2 \text{ and } y_1 \le y_c \le y_2,\\
0 & \text{otherwise.}
\end{cases}
\end{equation}

\noindent\textbf{Total Reward. }Combining all components, the final action-wise reward is computed as:

\begin{equation}
R = R_{\text{format}} \cdot w_1 + R_{\text{poin-in-box}} \cdot w_2,
\end{equation}
where $w_1$ and $w_2$ control the relative importance of format correctness and location precision.

\subsection{UI Navigation}

\subsubsection{Data Collection}

\begin{table*}[h]\small
\centering
\renewcommand{\arraystretch}{1.1}
\setlength{\tabcolsep}{6pt}
\begin{tabular}{c|cccccc|cc|c}
\toprule
\multirow{2}{*}{\textbf{Dataset}}           & \multicolumn{6}{c|}{\textbf{Multiple Steps}}                                                                                                                                                                         & \multicolumn{2}{c|}{\textbf{Single Step}}  & \multirow{2}{*}{\textbf{Sampled training set}}       \\ 
 & \multicolumn{1}{c}{{GUI-O}} & \multicolumn{1}{c}{{AITZ}} & \multicolumn{1}{c}{{AC}} & \multicolumn{1}{c}{{Aitw}}  & \multicolumn{1}{c}{{Amex}}  & \multicolumn{1}{c|}{{Navi*}}  & \multicolumn{1}{c}{{GD*}} & \multicolumn{1}{c|}{{VQA*}} &  \\ \hline
\textbf{\#Samples}     & \multicolumn{1}{c}{114k}      & \multicolumn{1}{c}{14k}  & \multicolumn{1}{c}{84k}           & \multicolumn{1}{c}{371k} & \multicolumn{1}{c}{39k} & \multicolumn{1}{c|}{20k}        & \multicolumn{1}{c}{107k}     & 42k & \textbf{350k}  \\ \
\textbf{\#Traces}    & \multicolumn{1}{c}{7k}        & \multicolumn{1}{c}{2k}   & \multicolumn{1}{c}{13k}           & \multicolumn{1}{c}{5k}  & \multicolumn{1}{c}{3k}  & \multicolumn{1}{c|}{*}          & \multicolumn{1}{c}{-}     & -  & * \\ \bottomrule
\end{tabular}
\caption{Basic statistics for our collected navigation training data. Note that ``AC'' means Android Control dataset while the ``GUI-O'' represents GUI-Odyssey. The datasets marked with * means are collected by ourselves. Among these data, we select 350k high-quality samples as our training set.}
\label{tab:navidata}
\end{table*}




For UI navigation tasks, we first collect traces from open-source navigation datasets of GUI Odyssey~\cite{guiodyssey}, Aguvis-Aitz~\cite{xu2024aguvis}, AndroidControl~\cite{androidcontrol}, Aitw~\cite{rawles2023androidinthewild}, and Amex~\cite{ominiact} as shown in Table~\ref{tab:navidata}. Meanwhile, grounding data plays a critical role in improving click operation accuracy for navigation tasks, so we also integrate grounding data into the training process. In addition, to strengthen the model's ability to handle long-chain traces and capabilities in cross-lingual generalization, we build a custom annotation platform that provides approximately 20k samples from several popular Chinese mobile APPs (notated Navi* in Table~\ref{tab:navidata}).  

Besides, a unified action space is critical for the integration of data from different sources, enabling the agent to focus on learning actions without being confused by diverse definitions. As shown in Table \ref{tab:action_space}, we follow UI-TARS but modify its action space (\emph{e.g.}, redefine \texttt{CallUser} action for information-retrieval) to better suit the existing open-source navigation training data.

\begin{table*}[ht]
\centering
\renewcommand{\arraystretch}{1.1} 
\scalebox{0.88}{
\begin{tabular}{ll}
\toprule
\textbf{Action} &\textbf{Definition}\\
\midrule
    Click(box=(x, y)) & Click at coordinates (x, y).\\
    Drag(start=(x1, y1), end=(x2, y2)) & Drag from (x1, y1) to (x2, y2).\\
    Scroll(start=(x1, y1), end=(x2, y2), direction=`') & Scroll from (x1, y1) to (x2, y2) with specified direction.\\
   Type(content=`')   & Type the specified content.\\
   Launch(app=`')   & Launch the specified app.\\
   Wait()   & Wait for loading.\\
   Finished(content=`')   & Finish the task, with optional information.\\
   CallUser(content=`')   & Conclude the answer for information-retrieval.\\
   LongPress(box=(x, y))    & Long press at coordinates (x, y).\\
   PressBack()    & Press the `back' button.\\
   PressHome()   & Press the `home' button.\\
   PressEnter() & Press the `enter' button.\\
   PressRecent() & Press the `recent' button.\\
\bottomrule
\end{tabular}
}
\caption{All actions and their definitions used in UI-Venus. We unify the action space and map all the actions in the existing open-source dataset to this space. }

\label{tab:action_space}
\end{table*}

\subsubsection{Data Pipeline}

As noisy data are detrimental to the training process of the model, we develop a pipeline to build a clean training set that can properly guide the model to learn various instructions, which includes three stages: (1) \textbf{Data Filtering} to reintegrate existing data, (2) \textbf{Trace Reconstruction} to modify existing traces, and (3) \textbf{Iteratively Trace Generation}
 to produce more high-quality traces beyond existing data. After data cleaning, we finally obtain about 350K samples for UI navigation training.

\noindent \textbf{Data Filtering.} In this stage, we preliminarily filter the collected data by removing overly short traces and standardizing the direction definition of scroll operations among different datasets. In addition, there are some inconsistencies between actions and tasks, \emph{e.g.}, there are traces that may contain more or less operations to complete tasks, or even not follow the requirements of tasks. To filter out these invalid traces, we utilize MLLM to summarize each action and integrate the summaries to obtain an overall description for each trace, which could be compared with the original task. Finally, we categorize the traces based on the apps and subtasks they involve and resample them to ensure the diversity of the training data, preventing the model from overfitting to specific scenarios without sufficient exploration on diverse tasks.

\noindent \textbf{Trace Reconstruction.} Among existing datasets, the information retrieval tasks~\cite{androidworld}, an important category of UI navigation tasks, often lack an explicit answer in the final step of the traces.
For example, when users make a request of `What is the weather today?' or `What is the total price in my shopping cart now?', they usually expect an answer from the agent, rather than just being navigated to some pages and finding the answer by themselves. To bridge this gap, we select traces of information retrieval tasks from filtered data, and then adopt the MLLM to generate corresponding answers based on the last screenshots of these traces. Finally, we reconstruct the original trace by inserting a \texttt{CallUser} step with generated answer before the \texttt{Finish} step, requiring the agent to report the final answer before ending an information retrieval task.

\noindent \textbf{Iteratively Trace Generation.} Besides utilizing existing open-source trajectory data, we also design an automated framework to iteratively generate high-quality traces built upon the virtual cloud environment, which contains dozens of available mobiles. We adopt our well-trained UI-Venus model to generate trajectories on real-world tasks including hand-designed and MLLM-generated instructions to ensure their diversity in Chinese and English mobile applications. The noisy and invalid traces would be discarded by combining following steps: (1) \textbf{Rule-based filtering}: empirical rules are defined to remove typical errors (\emph{e.g.}, short trajectory with abnomal exit, the repeating invalid actions); (2) \textbf{ORM-based filtering}: we trained an outcome reward model (ORM) regarding the whole trajectory as a single example to score on the generated traces and remove those with low scores; (3) \textbf{Annotator-based filtering}: Annotators are required to strictly select correct traces from the remaining ones. Also, to learn from failures, the fault actions would be fixed by annotators and added to the train set with its valid trace prefix. The above process helps iteratively optimize our model for complex real-world scenarios.

\subsubsection{Trajectory History Alignment and Sparse Action Enhancement}

Accurate historical context is critical for successful task execution in dynamic UI navigation planning. By leveraging historical information, the agent model can understand the steps already taken, perceive the state transitions~\cite{qin2025uitarspioneeringautomatedgui,chae2025web,sun2025guixploreempoweringgeneralizablegui}, and engage in self-reflection~\cite{shinn2023reflexion,GUI_Reflection,li2025mobileuse}.

In our approach, the historical context is provided as thought-action pairs, where the thought represents the reasoning process behind taking a particular action at each step, and the action is the actual executed operation. Specifically, when predicting the $n$-th step, historical context from the previous $n-1$ steps is formulated as:

\begin{equation}
H_{n-1} = \big[ (t_1, a_1), (t_2, a_2), \dots, (t_{n-1}, a_{n-1}) \big].
\end{equation}

The context $H_{n-1}$, together with the task description and the current UI screenshot, forms the input to the agent model for the $n$-th step prediction.

However, in practice, the historical thoughts $(t_1, \dots, t_{n-1})$ are often not well aligned with the agent model’s intrinsic reasoning capability~\cite{li2025reasoningaslogicunitsscalingtesttimereasoning}, due to inconsistencies in style, detail, and abstraction level across data. This mismatch reduces the utility of $H_{n-1}$, sometimes causing confusion in decision-making. 

To address this issue, we propose a self-evolving~\cite{tao2024surveyselfevolutionlargelanguage} history alignment mechanism that refines historical thoughts between training epochs, progressively aligning the reasoning traces (\emph{i.e.}, the thought sequence in $H_{n-1}$) with the model’s evolving decision patterns. By dynamically adjusting the trajectory history to better reflect the model’s current reasoning behavior, the mechanism provides a more coherent and consistent historical context, resulting in sustained improvements in navigation performance.

Additionally, we observe that the action distribution in the training data significantly impacts the model performance~\cite{qi2025webrltrainingllmweb}. Particularly for sparse actions (\emph{e.g.},~\texttt{LongPress}), the agent model often struggles to learn these actions effectively due to their low frequency in the data. In certain complex trajectories, these actions can be pivotal, and missing them may cause the entire task to fail. To overcome the limitation, our method also enhances the sampling of sparse actions, allowing the agent to better acquire and generalize these rare but critical skills. The sampling strategy helps the model develop a more transferable understanding of both the UI states and the planned actions, which improves its robustness in complex and dynamic task scenarios.

\begin{figure}[h]
\centering
\includegraphics[width=1.0\linewidth]{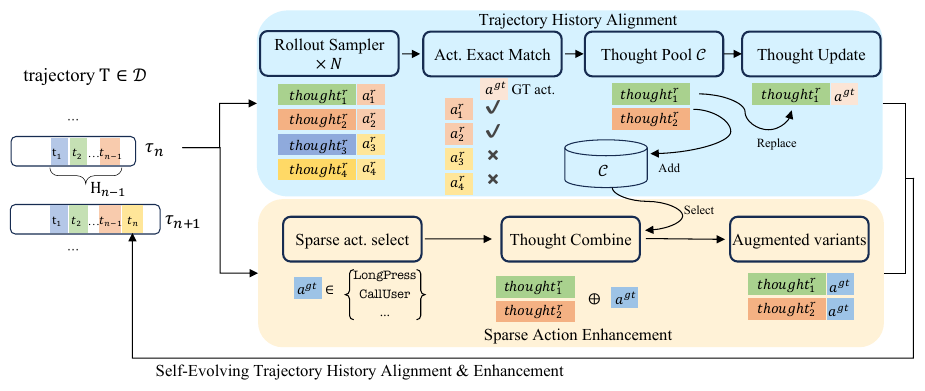}
\caption{
The overview of the proposed Self-Evolving Trajectory History Alignment \& Enhancement process, applied between training epochs. The process consists of two key components: 1) \textbf{Trajectory History Alignment} refines the historical context for each trajectory step. The model executes multiple rollouts to generate candidate thought–action pairs, applying an Action Exact Match filter to retain only those whose predicted actions match ground-truth actions. The corresponding thoughts are collected in pool $\mathcal{C}$ and subsequently replace the original thoughts in the historical context, creating an optimized trajectory history for the next training epoch. 2) \textbf{Sparse Action Enhancement} focuses on samples with sparse actions. Multiple variants are constructed by combining different rollout generated thoughts that lead to the same sparse action, effectively increasing the representation of these sparse but critical operations in the training distribution.
}
\label{fig:history_align_overall}
\end{figure}

To elaborate on these enhancements, we now provide a detailed description of the two key components of our approach. Figure~\ref{fig:history_align_overall} illustrates the overall framework of the proposed \textit{Self-Evolving} Trajectory History Alignment \& Enhancement method.

\noindent \textbf{Trajectory History Alignment. }
In our framework, the agent model leverages historical context composed of thought–action pairs from previous steps, where each thought describes the reasoning process behind the corresponding action. Explicitly incorporating the past chain-of-thought (CoT)~\cite{wei2023chainofthoughtpromptingelicitsreasoning} in this historical context can improve planning quality in sequential decision-making tasks by revealing the latent reasoning behind each action. However, existing UI navigation datasets with high-quality CoT labels are both scarce and heterogeneous. These annotations are collected from diverse sources, including crowd-sourced workers~\cite{rawles2023androidwildlargescaledataset}, expert demonstrations~\cite{zhang2024androidzoochainofactionthoughtgui}, and synthetic generation~\cite{sun2025osgenesisautomatingguiagent}, which leads to substantial variation in sequence length, linguistic style, and level of reasoning detail.

As historical thoughts play a crucial role in guiding the agent model’s planning, cross-source variability can introduce inconsistencies in its perception for the navigation history, ultimately degrading decision quality. In particular, shorter thought sequences may miss critical intermediate observations, while overly verbose ones may obscure key decision-making milestones.

To address these challenges, we introduce a self-evolving trajectory history alignment mechanism that iteratively refines historical thoughts during training. The detailed procedure is presented in Algorithm~\ref{alg:history-alignment}.

After each training epoch, we perform global trajectory refinement by re-inferring reasoning traces with the current model. Let $\mathcal{D}=\{T_k\}_{k=1}^K$ denote the dataset. 
Each trajectory $T_k$ is a sequence of steps 
$\{(x_{k,n}, H_{k,n-1})\}_{n=1}^{N_k}$, where 
$x_{k,n}$ is the UI state (screenshot) at step $n$, 
$H_{k,n-1}=\big[(t_{k,1},a_{k,1}),\dots,(t_{k,n-1},a_{k,n-1})\big]$ 
is the historical context from the previous $n-1$ steps.

For each step $n$, given $(x_{k,n}, H_{k,n-1})$, we perform $R$ rollouts:
\begin{equation}
\{(t^{(r)}_{k,n},a^{(r)}_{k,n})\}_{r=1}^R \sim p_{\theta}(t,a\,|\,x_{k,n},H_{k,n-1}).
\end{equation}
We then filter the results to keep only those whose predicted action matches the ground-truth action:
\begin{equation}
\mathcal{C}_{k,n}=\{\,t^{(r)}_{k,n} \mid a^{(r)}_{k,n} = a^{(gt)}_{k,n} \,\}.
\end{equation}
All candidates in $\mathcal{C}_{k,n}$ form the thought pool $\mathcal{C}$.

For the $n$-th step, the thoughts in all previous $n-1$ steps are refined by selecting a replacement from the corresponding thought pool:
\begin{equation}
t'_{k,i} = 
\begin{cases}
\text{Select}(\mathcal{C}_{k,i}), & \mathcal{C}_{k,i} \neq \varnothing, \\
t_{k,i}, & \text{otherwise},
\end{cases}
\quad i = 1, \dots, n-1,
\end{equation}
where $\text{Select}(\cdot)$ denotes a selection policy (\emph{e.g.}, choosing the candidate with length closest to a target length). 

The updated historical context $H'_{k,n-1} = \big[(t'_{k,1},a^{(gt)}_{k,1}), \dots, (t'_{k,n-1},a^{(gt)}_{k,n-1})\big]$ is then used as input in the next training epoch. 
This refinement is applied after every epoch, aligning the reasoning traces with the agent’s evolving policy. This self-evolving process enhances the agent model’s planning, leading to more robust and coherent navigation performance.

\begin{algorithm}[t]
\caption{Self-Evolving Trajectory History Alignment}
\label{alg:history-alignment}
\begin{algorithmic}[1]
\Require Dataset $\mathcal{D}=\{T_k\}_{k=1}^K$; each trajectory $T_k$ is a sequence of steps $\{(x_{k,n}, H_{k,n-1})\}_{n=1}^{N_k}$; ground-truth actions $\{a_{k,1:N_k}\}$; model $p_\theta(t,a\,|\,x,H)$; rollout count $R$
\Ensure Updated historical thoughts $\{t_{k,1:N_k}\}$
\For{each training epoch}
  \For{each trajectory $T_k$}
    \State Initialize thought pools $\{\mathcal{C}_{k,i}\leftarrow \emptyset\}_{i=1}^{N_k}$
    \For{$n=1$ to $N_k$} \Comment{collect candidates for every step}
      \State $H_{k,n-1}\leftarrow [(t_{k,1},a_{k,1}),\dots,(t_{k,n-1},a_{k,n-1})]$
      \For{$r=1$ to $R$} \Comment{rollouts for step $n$}
        \State $(t^{(r)}_{k,n}, a^{(r)}_{k,n}) \sim p_\theta(t,a\,|\,x_{k,n}, H_{k,n-1})$
        \If{$a^{(r)}_{k,n} = a^{(gt)}_{k,n}$} \Comment{action exact-match}
          \State $\mathcal{C}_{k,n}\leftarrow \mathcal{C}_{k,n}\cup\{t^{(r)}_{k,n}\}$ \Comment{add to thought pool}
        \EndIf
      \EndFor
    \EndFor
    \For{$i=1$ to $N_k$} \Comment{select historical thoughts after collecting pools}
      \If{$\mathcal{C}_{k,i}\neq \emptyset$}
        \State $t'_{k,i}\leftarrow \textsc{Select}(\mathcal{C}_{k,i})$ \Comment{select policy}
      \EndIf
    \EndFor
    \For{$n=1$ to $N_k$} \Comment{update historical context for next epoch}
      \State $H_{k,n-1}\leftarrow [(t'_{k,1},a^{(gt)}_{k,1}),\dots,(t'_{k,n-1},a^{(gt)}_{k,n-1})]$
    \EndFor
  \EndFor
\EndFor
\end{algorithmic}
\end{algorithm}

\begin{figure}[h]
\centering
\includegraphics[width=0.75\linewidth]{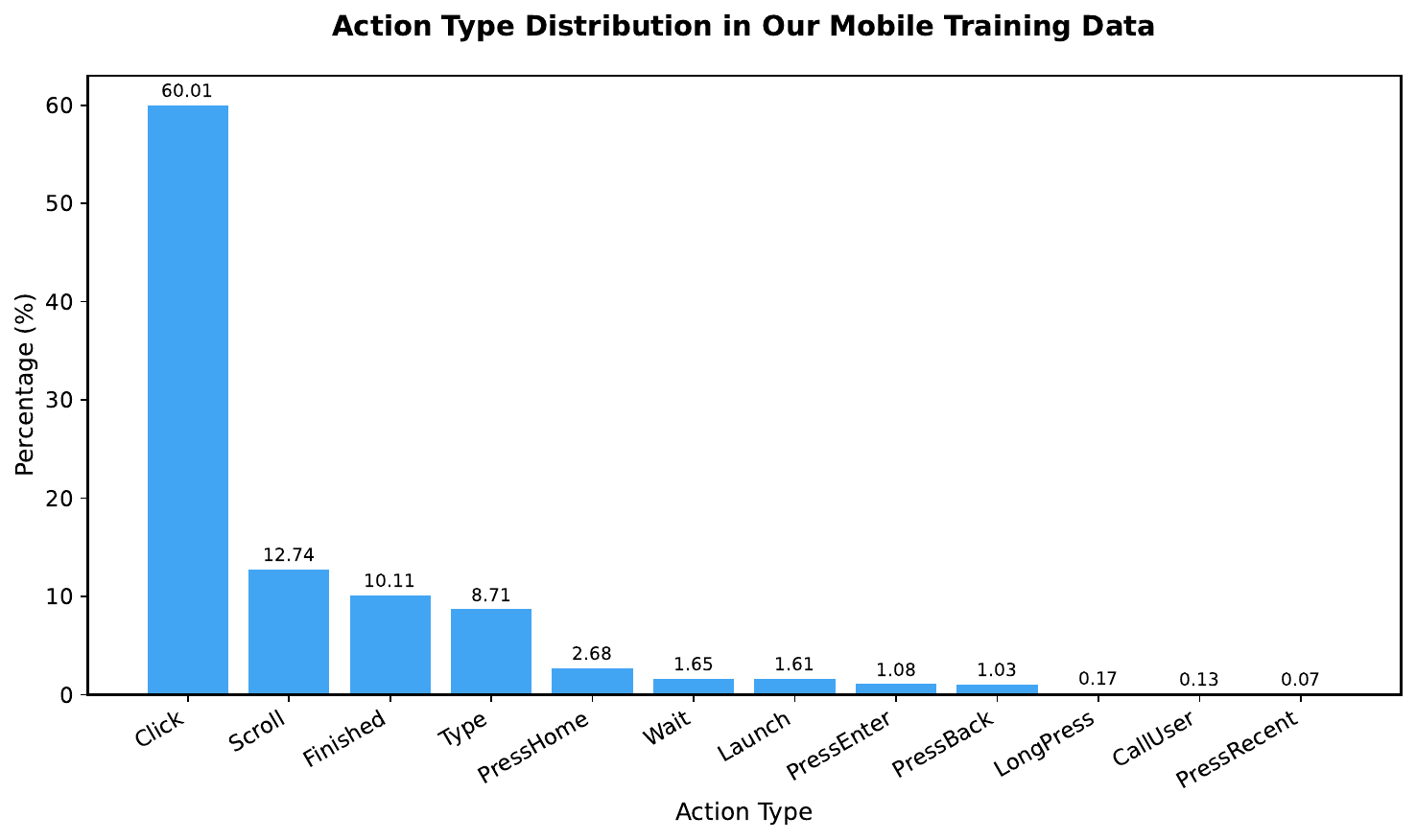}
\caption{
Action type distribution in our mobile training data, showing a long-tailed profile with several low-frequency (sparse) actions.}
\label{fig:action_distri}
\end{figure}

\noindent \textbf{Sparse Action Enhancement. }
The distribution of actions in the training data is often imbalanced (see Figure~\ref{fig:action_distri}). Common actions such as \texttt{Click} and \texttt{Scroll} appear frequently, while rare actions such as \texttt{LongPress} are much less represented. This imbalance makes it difficult for the model to learn sparse actions effectively. However, these sparse actions often play a critical role in complex task completion, and insufficient‌ handling of such actions results in incomplete or erroneous navigation plans.

To address this problem, we design a sampling strategy that increases learning for sparse actions. 
During trajectory history alignment, the agent produces a thought pool $\mathcal{C}_n$ at each step via rollouts. 
For a step $(x_n, H_{n-1})$ involving a sparse action (\emph{e.g.}, $a_n\in\{\text{LongPress},\text{CallUser},\dots\}$), we create $M$ historical variants by combining thoughts from the pools. 
Specifically, for each previous step $i=1,\dots,n-1$, we sample $\{t’^{(m)}_{i}\}_{m=1}^{M} \subset \mathcal{C}_i$ and construct $H^{(m)}_{n-1}=\big[(t'^{(m)}_{1},a_{1}),\dots,(t'^{(m)}_{n-1},a_{n-1})\big]$. 
Conceptually, the augmented set corresponds to the Cartesian product $\mathcal{C}_1 \times \cdots \times \mathcal{C}_{n-1}$. This increases both the diversity and frequency of reasoning patterns that lead to sparse actions. For common actions, we do not perform this augmentation. By focusing only on sparse actions, we enhance the model’s ability to learn these rare behaviors, leading to more accurate navigation reasoning.

\subsubsection{Action-wise Reward Function}

The rule-based reward function is a key component in reinforcement learning, especially for tasks where outputs can be verified against well-defined rules. In visual grounding, for example, the reward is often based on the intersection-over-union between the predicted and the ground-truth bounding boxes~\cite{shen2025vlmr1stablegeneralizabler1style}. While effective for coarse-level correctness, such rewards provide limited guidance for fine-grained UI manipulation.

In UI navigation tasks, the agent model is required to select the correct action type (\emph{e.g.}, \texttt{Click}, \texttt{Scroll}) and generate precise action parameters (\emph{e.g.}, target coordinates or input text). This requirement becomes particularly important in complex interfaces, where even small parameter errors can cause task failure. We propose an \textit{action-wise reward function} tailored for the navigation task, which assesses the output action along multiple dimensions, namely format, action type, coordinate, and content, to provide detailed feedback. It consists of four components:

\noindent\textbf{Format Reward. }
We apply a format reward to train the model to generate reasoning and action outputs that follow a predefined template. This design is motivated by the fact that the reasoning process provides useful historical context for subsequent decision-making. The format reward, denoted as $R_{\mathrm{format}}$, checks whether the model output contains the required XML-style tags in the correct order, with the reasoning block enclosed in \texttt{<think>} followed by the action block enclosed in \texttt{<action>}~\cite{deepseekai2025deepseekr1incentivizingreasoningcapability,qian2025toolrlrewardtoollearning}.

\noindent\textbf{Action Type Reward. }
We adopt the action type reward, $R_{\text{type}}$, which compares the predicted action type with the ground-truth action type. The model receives a reward of 1 for a match and 0 otherwise. This provides a direct and interpretable metric to encourage accurate action type prediction~\cite{lu2025uir1enhancingefficientaction}.

\noindent\textbf{Coordinate Reward. }
For actions involving spatial positioning (\emph{e.g.}, \texttt{Click}, \texttt{Scroll}), we adopt a coordinate reward, $R_{\text{coord}}$, that depends on the pixel-level distance between the predicted and ground-truth coordinates. Prior methods often assess correctness by checking whether this distance is within a fixed threshold, which can be somewhat rigid and insufficient for distinguishing near-misses from large deviations. To address this, we introduce a stepwise reward strategy that grants higher scores for predictions closer to the target, while still awarding partial credit for reasonably accurate locations. This design encourages the model to incrementally improve its coordinate parameter localization.





For point-targeting actions (\emph{i.e.}, \texttt{Click}, \texttt{LongPress}), we adopt a stepwise coordinate reward based on the pixel-level distance $d$ between the predicted and the ground-truth coordinates:
\begin{equation}
R_{\text{coord}} =
\begin{cases}
\alpha, & d < \delta_2, \\
0.5\alpha, & \delta_2 \leq d < \delta_1, \\
0, & \text{otherwise}.
\end{cases}
\end{equation}
Here, $\delta_1$ and $\delta_2$ ($\delta_2 < \delta_1$) are distance thresholds that assign higher scores to more accurate predictions, while still granting partial credit for reasonably close locations.

For scrolling actions, we incorporate both spatial accuracy at the start and end positions as well as the correctness of the scrolling direction. The reward $R_{\text{scroll}}$ is defined as:
\begin{equation}
R_{\text{scroll}} =
\begin{cases}
1.5\beta & \text{if } d_{\text{start}}, d_{\text{end}} < \delta_3 \text{ and direction match}, \\
\beta & \text{if } d_{\text{start}} < \delta_3 \text{ and direction match}, \\
0.5\beta & \text{if } d_{\text{start}} < \delta_3 \text{ or direction match}, \\
0 & \text{otherwise}.
\end{cases}
\end{equation}
Here, $d_{\text{start}}$ and $d_{\text{end}}$ denote the pixel-level distances between the predicted and ground-truth start and end coordinates, respectively, and $\delta_3$ is the spatial distance threshold for scroll evaluation. The direction refers to the scrolling orientation (\emph{i.e.}, \texttt{up}, \texttt{down}, \texttt{left}, or \texttt{right}), and a direction match indicates that the predicted orientation exactly matches the ground-truth.

\noindent\textbf{Content Reward. }
For actions involving text input (\emph{e.g.}, \texttt{Type}), we compute the reward $R_{\text{content}}$ using the token-level F1-score between the predicted and ground-truth text:
\begin{equation}
R_{\text{content}} =
\begin{cases}
\gamma & \text{if F1-score} \ge 0.5, \\
0 & \text{otherwise}.
\end{cases}
\end{equation}
Here, the F1 score captures the precision-recall balance of textual overlap between the prediction and the reference.

\noindent\textbf{Total Reward. }
Combining all components, the final action-wise reward is computed as:

\begin{equation}
R = R_{\text{format}} \cdot w_1 + (R_{\text{type}} + R_{\text{coord}} + R_{\text{content}}) \cdot w_2,
\end{equation}
where $w_1$ and $w_2$ control the relative importance of structural correctness and action parameter precision.



\section{Experiments}

\subsection{Implementation Details}
We summarize the training hyperparameters in Table~\ref{tab:hyper} and provide the implementation details in the following section.

\begin{table}[ht]\small
\centering
\begin{tabular}{lccccc}
\toprule
\textbf{Hyperparameter} & \textbf{rollout} & \textbf{batch\_size} & \textbf{$\beta_{KL}$}  & \textbf{freeze\_vit} & \textbf{epoch}\\
\midrule
    \textbf{UI-Venus-Ground-7B}  & 8  & 128 & 4e-3 & False & 10  \\
    \textbf{UI-Venus-Ground-72B} & 10 & 128 & 4e-3 & False & 1  \\
    \textbf{UI-Venus-Navi-7B}    & 8  & 256 & 1e-3 & True  & 10  \\
   \textbf{UI-Venus-Navi-72B}    & 8  & 512 & 1e-3 & True  & 1 \\
\bottomrule
\end{tabular}
\vspace{0.4em}
\caption{Training hyperparameter settings used in the experiments of UI-Venus.}
\label{tab:hyper}
\end{table}
\begin{table}[ht]
    \centering
    \footnotesize
    \setlength{\tabcolsep}{0pt}
    \begin{tabular*}{\columnwidth}{@{\extracolsep{\fill}}l *{7}{c}}
    \toprule
    \multirow{2}{*}{\textbf{Models}} & \multicolumn{2}{c}{\textbf{Mobile}}  & \multicolumn{2}{c}{\textbf{Desktop}} & \multicolumn{2}{c}{\textbf{Web}}     & \multirow{2}{*}{\textbf{Avg}} \\ 
    \cmidrule(lr){2-3} \cmidrule(lr){4-5} \cmidrule(lr){6-7}
                                     & \textbf{Text} & \textbf{Icon/Widget} & \textbf{Text} & \textbf{Icon/Widget} & \textbf{Text} & \textbf{Icon/Widget} &                               \\ 
    \midrule
    \rowcolor{gray!15}
    \multicolumn{8}{l}{\textit{Closed-source Models}} \\
    GPT-4o~\citep{openai2024gpt4o}                           & 26.6          & 24.2                 & 24.2          & 19.3                 & 12.8          & 11.8                 & 20.1                          \\ 
    UI-TARS-1.5~\citep{ui-tars-15-seed}                           & -          & -                 & -          & -                 & -          & -                 & 94.2                          \\
    Seed1.5-VL~\citep{seed2025seed1_5vl}                           & -          & -                 & -          & -                & -          & -                 & {\ul95.2}                          \\
    \midrule
    \rowcolor{gray!15}
    \multicolumn{8}{l}{\textit{General Open-source Models}} \\
    Qwen2.5-VL-7B*~\citep{bai2025qwen25vltechnicalreport}                      & 98.3          & 85.3                & 88.7          & 58.6                 & 92.7          & 81.8                 & 87.7                         \\ 
    Qwen2.5-VL-72B*~\citep{bai2025qwen25vltechnicalreport}                      & 97.6          & 88.6                 & 92.3          & 86.6                 & 91.9          & 85.2                 & 90.7                          \\ 
    \midrule
    \rowcolor{gray!15}
    \multicolumn{8}{l}{\textit{GUI-specific Models (SFT)}} \\
    SeeClick-9.6B~\citep{cheng2024seeclickharnessingguigrounding}                    & 78.4          & 50.7                 & 70.1          & 29.3                 & 55.2          & 32.5                 & 55.1                          \\
    ShowUI-2B~\cite{lin2024showuivisionlanguageactionmodelgui}                        & 92.1          & 75.4                 & 78.9          & 78.9                 & 84.2          & 61.1                 & 77.3                          \\
    UGround-7B~\citep{gou2024navigatingdigitalworldhumans}                       & 75.1          & 84.5                 & 85.1          & 61.4                 & 84.6          & 71.9                 & 76.3                          \\
    OS-Atlas-7B~\citep{wu2024osatlasfoundationactionmodel}                 & 95.2          & 75.8                 & 90.7          & 63.6                 & 90.6          & 77.3                 & 84.1                          \\
    Aguvis-7B~\citep{xu2024aguvis}                        & 89.3          & 68.7                 & 80.6          & 67.9                 & 89.3          & 70.0                 & 80.5                          \\
    UI-TARS-7B~\citep{qin2025uitarspioneeringautomatedgui}                        & 96.9          & 89.1                 & 95.4          & 85.0                 & 93.6          & 85.2                 & 91.6                          \\
    UI-TARS-72B~\citep{qin2025uitarspioneeringautomatedgui}                       & 94.8          & 86.3                 & 91.2          & 87.9                 & 91.5          & 87.7                 & 90.3                          \\
    \textsc{Jedi}-7B~\citep{xie2025scalingcomputerusegroundinguser}                          & 96.9          & 87.2                 & 95.9          & 87.9                 & 94.4          & 84.2                 & 91.7                          \\  
    GUI-Actor-7B~\citep{guiactor}                    & 97.6          & 88.2                 & 96.9          & 85.7                 & 93.2          & 86.7                 & 92.1                          \\
    OpenCUA-7B~\citep{wang2025opencuaopenfoundationscomputeruse} & -          & -                  & -           & -                  & -           & -                  & 92.3                          \\
    OpenCUA-32B~\citep{wang2025opencuaopenfoundationscomputeruse} & -           & -                  & -           & -                  & -           & -                  & 93.4                          \\
    \midrule
    \rowcolor{gray!15}
    \multicolumn{8}{l}{\textit{GUI-specific Models (RL)}} \\
    UI-R1-E-3B~\citep{lu2025uir1enhancingefficientaction}                       & 98.2          & 83.9                 & 94.8          & 75.0                 & 93.2          & 83.7                 & 89.5                          \\
    SE-GUI-7B~\citep{yuan2025enhancingvisualgroundinggui}                        & -             & -                    & -             & -                    & -             & -                    & 90.3                          \\
    LPO~\citep{tang2025lpoaccurateguiagent}                              & 97.9          & 82.9                 & 95.9          & 86.4                 & { \ul 95.6}          & 84.2                 & 90.5                          \\
    GUI-G$^2$-7B~\citep{guig2}                                           & -          & -                 & -          & -                 & -          & -                 & 93.3 \\
    Phi-Ground-7B-16C-DPO~\citep{phiground}                       & 96.5          & 62.0       & 90.2          & 76.4                 & 93.6          & 75.9               &  83.8                        \\
    GTA1-7B$\dagger$~\citep{yang2025gta1guitesttimescaling}  & 99.0          & 88.6                 & 94.9          & 89.3                & 92.3          & 86.7                 & 92.4                         \\
    GTA1-72B~\citep{yang2025gta1guitesttimescaling}  & {\ul 99.3}          & {\ul 92.4}                 & \textbf{97.4}          & 89.3                & 95.3          & {\ul 91.4 }               & 94.8                          \\
    \midrule
    \rowcolor{gray!15}
    \multicolumn{8}{l}{\textit{Ours}} \\
    UI-Venus-Ground-7B                        & 99.0  & 90.0         & {\ul 97.0}  & \textbf{90.7} & \textbf{96.2} & 88.7       &      94.1               \\ 
    UI-Venus-Ground-72B   & \textbf{99.7}  & \textbf{93.8}        & 95.9  & {\ul 90.0} & \textbf{96.2} & \textbf{92.6}         & \textbf{95.3}      \\ 
    \bottomrule
    \end{tabular*}
    \vspace{0.4em}
    \caption{Performance comparison on \textbf{ScreenSpot-V2} dataset. Our UI-Venus-72B achieves state-of-the-art performance, outperforming all baseline methods across mobile, desktop, and web platforms. Note that models with * are reproduced and the $\dagger$ means trained from UI-TARS-1.5-7B.}
    \label{tab:screenspot-v2}
\end{table}

\noindent \textbf{Grounding.} For UI grounding, we train our model with the above-mentioned clean data based on Qwen2.5-VL~\cite{bai2025qwen25vltechnicalreport}. Specifically, we set the learning rate as $4\times 10^{-7}$ and the rollout sample $8$, $10$ for 7B, 72B model, respectively. The global batch size is $128$ and the models are trained with 128 ppu-gpus with the EasyR1 framework~\cite{easyr1}. It takes 1.5 days, 10 days to train a 7B, 72B model for about one epoch, respectively. 
To fully reproduce our benchmark results, we provide our evaluation prompt for grounding in Appendix \ref{sec:gd_prompt}.

\noindent \textbf{Navigation.} Based on the filtered navigation training data, the training is conducted with a learning rate of $4\times 10^{-7}$ and a rollout sample size of $8$. The global batch size is set to $256$ and $512$ for the 7B and 72B models, respectively, with training conducted on 256 and 512 PPU-GPUs. Training one epoch takes approximately 1 day for the 7B model and 8.5 days for the 72B model. In addition, we provide the user prompt for the navigation task in Appendix~\ref{sec:navi_prompt}, where the model is constrained to output its reasoning and conclusions along with the predicted actions in a specified format.




\subsection{Grounding Benchmarks}

\begin{table*}[t]
    \centering
    \footnotesize
    \setlength{\tabcolsep}{0pt}
    \begin{tabular*}{\textwidth}{@{\extracolsep{\fill}}l *{15}{c}}
    \toprule
    \multirow{3}{*}{\textbf{Model}} & \multicolumn{2}{c}{\textbf{CAD}} & \multicolumn{2}{c}{\textbf{Dev}} & \multicolumn{2}{c}{\textbf{Creative}} & \multicolumn{2}{c}{\textbf{Scientific}} & \multicolumn{2}{c}{\textbf{Office}} & \multicolumn{2}{c}{\textbf{OS}} & \multicolumn{3}{c}{\textbf{Avg.}} \\
    \cmidrule(lr){2-3} \cmidrule(lr){4-5} \cmidrule(lr){6-7} \cmidrule(lr){8-9} \cmidrule(lr){10-11} \cmidrule(lr){12-13} \cmidrule(lr){14-16}
    & Text & Icon & Text & Icon & Text & Icon & Text & Icon & Text & Icon & Text & Icon & Text & Icon & \textbf{Avg.} \\
    \midrule
    \rowcolor{gray!15}
    \multicolumn{16}{l}{\textit{Closed-source Models}} \\
    GPT-4o~\citep{openai2024gpt4o} & 2.0 & 0.0 & 1.3 & 0.0 & 1.0 & 0.0 & 2.1 & 0.0 & 1.1 & 0.0 & 0.0 & 0.0 & 1.3 & 0.0 & 0.8 \\
    Claude Computer Use~\citep{anthropic2024cuda} & 14.5 & 3.7 & 22.0 & 3.9 & 25.9 & 3.4 & 33.9 & 15.8 & 30.1 & 16.3 & 11.0 & 4.5 & 23.4 & 7.1 & 17.1 \\
    UI-TARS-1.5~\citep{ui-tars-15-seed} & - & - & - & - & - & - & - & - & - & - & - & - & - & - & {\ul 61.6} \\
    Seed1.5-VL~\citep{seed2025seed1_5vl} & - & - & - & - & - & - & - & - & - & - & - & - & - & - & 60.9 \\
    \midrule
    \rowcolor{gray!15}
    \multicolumn{16}{l}{\textit{General Open-source Models}} \\
    Qwen2.5-VL-7B*~\citep{bai2025qwen25vltechnicalreport} & 16.8 & 1.6 & 46.8 & 4.1 & 35.9 & 7.7 & 49.3 & 7.3 & 52.5 & 20.8 & 37.4 & 6.7 & 38.9 & 7.1 & 26.8 \\
    Qwen2.5-VL-72B*~\citep{bai2025qwen25vltechnicalreport} & 54.8 & 15.6 & 65.6 & 16.6 & 63.1 & 19.6 & 78.5 & 34.5 & 79.1 & 47.2 & 66.4 & 29.2 & 67.3 & 25.0 & 51.2 \\
    \midrule
    \rowcolor{gray!15}
    \multicolumn{16}{l}{\textit{GUI-specific Models (SFT)}} \\
    SeeClick-9.6B~\citep{cheng2024seeclickharnessingguigrounding} & 2.5 & 0.0 & 0.6 & 0.0 & 1.0 & 0.0 & 3.5 & 0.0 & 1.1 & 0.0 & 2.8 & 0.0 & 1.8 & 0.0 & 1.1 \\
    \textsc{Focus}-2B~\citep{tang2025thinktwiceclickonce} & 7.6 & 3.1 & 22.8 & 1.7 & 23.7 & 1.7 & 25.0 & 7.1 & 23.2 & 7.7 & 17.8 & 2.5 & 19.8 & 3.9 & 13.3 \\
    CogAgent-18B~\citep{hong2024cogagentvisuallanguagemodel} & 7.1 & 3.1 & 14.9 & 0.7 & 9.6 & 0.0 & 22.2 & 1.8 & 13.0 & 0.0 & 5.6 & 0.0 & 12.0 & 0.8 & 7.7 \\
    Aria-UI~\citep{yang2024ariauivisualgroundinggui} & 7.6 & 1.6 & 16.2 & 0.0 & 23.7 & 2.1 & 27.1 & 6.4 & 20.3 & 1.9 & 4.7 & 0.0 & 17.1 & 2.0 & 11.3 \\
    OS-Atlas-7B~\citep{wu2024osatlasfoundationactionmodel} & 12.2 & 4.7 & 33.1 & 1.4 & 28.8 & 2.8 & 37.5 & 7.3 & 33.9 & 5.7 & 27.1 & 4.5 & 28.1 & 4.0 & 18.9 \\
    ShowUI-2B~\citep{lin2024showuivisionlanguageactionmodelgui} & 2.5 & 0.0 & 16.9 & 1.4 & 9.1 & 0.0 & 13.2 & 7.3 & 15.3 & 7.5 & 10.3 & 2.2 & 10.8 & 2.6 & 7.7 \\
    UGround-7B \citep{gou2024navigatingdigitalworldhumans} & 14.2 & 1.6 & 26.6 & 2.1 & 27.3 & 2.8 & 31.9 & 2.7 & 31.6 & 11.3 & 17.8 & 0.0 & 25.0 & 2.8 & 16.5 \\
    UGround-V1-7B~\citep{gou2024navigatingdigitalworldhumans} & 15.8 & 1.2 & 51.9 & 2.8 & 47.5 & 9.7 & 57.6 & 14.5 & 60.5 & 13.2 & 38.3 & 7.9 & 45.2 & 8.1 & 31.1 \\
    UI-TARS-7B~\citep{qin2025uitarspioneeringautomatedgui} & 20.8 & 9.4 & 58.4 & 12.4 & 50.0 & 9.1 & 63.9 & 31.8 & 63.3 & 20.8 & 30.8 & 16.9 & 47.8 & 16.2 & 35.7 \\
    UI-TARS-72B~\citep{qin2025uitarspioneeringautomatedgui}& 18.8 & 12.5 & 62.9 & 17.2 & 57.1 & 15.4 & 64.6 & 20.9 & 63.3 & 26.4 & 42.1 & 15.7 & 50.9 & 17.6 & 38.1 \\
    \textsc{Jedi}-7B~\citep{xie2025scalingcomputerusegroundinguser} & 38.0 & 14.1 & 42.9 & 11.0 & 50.0 & 11.9 & 72.9 & 25.5 & 75.1 & 47.2 & 33.6 & 16.9 & 52.6 & 18.2 & 39.5 \\
    GUI-Actor-7B~\citep{guiactor} & - & - & - & - & - & - & - & - & - & - & - & - & - & - & 44.6 \\
    OpenCUA-7B~\citep{wang2025opencuaopenfoundationscomputeruse}& - & - & - & - & - & - & - & - & - & - & - & - & - & - & 50.0 \\
    OpenCUA-32B~\citep{wang2025opencuaopenfoundationscomputeruse}& - & - & - & - & - & - & - & - & - & - & - & - & - & - & 55.3 \\
    \midrule
    \rowcolor{gray!15}
    \multicolumn{16}{l}{\textit{GUI-specific Models (RL)}} \\
    UI-R1-E-3B~\citep{lu2025uir1enhancingefficientaction} & 37.1 & 12.5 & 46.1 & 6.9 & 41.9 & 4.2 & 56.9 & 21.8 & 65.0 & 26.4 & 32.7 & 10.1 & - & - & 33.5 \\
    GUI-R1-7B~\citep{luo2025guir1generalistr1style} & 23.9 & 6.3 & 49.4 & {4.8} & 38.9 & {8.4} & 55.6 & 11.8 & 58.7 & {26.4} & {42.1} & {16.9} & - & - & - \\
    InfiGUI-R1-3B~\citep{liu2025infiguir1advancingmultimodalgui} & 33.0 & 14.1 & 51.3 & 12.4 & 44.9 & 7.0 & 58.3 & 20.0 & 65.5 & 28.3 & 43.9 & 12.4 & 49.1 & 14.1 & 35.7 \\ 
    GUI-G1-3B~\citep{zhou2025guig1understandingr1zeroliketraining} & 39.6 & 9.4 & 50.7 & 10.3 & 36.6 & 11.9 & 61.8 & 30.0 & 67.2 & 32.1 & 23.5 & 10.6 & 49.5 & 16.8 & 37.1 \\
    SE-GUI-7B~\citep{yuan2025enhancingvisualgroundinggui} & 51.3 & \textbf{42.2} & 68.2 & {19.3} & {57.6} & 9.1 & 75.0 & {28.2} & {78.5} & {43.4} & 49.5 & {25.8} & 63.5 & {21.0} & 47.3 \\
    Phi-Ground-7B-16C-DPO~\citep{phiground} & 26.9  & 17.2 & 70.8  & 16.7 & 56.6 & 13.3 & 58.0  & 29.1  & 76.4  & 44.0 & 55.1 & 25.8 & 56.4 & 21.8 & 43.2 \\
    GUI-G$^2$-7B~\citep{guig2} & 55.8 & 12.5 & 68.8 & 17.2 & 57.1 & 15.4 & 77.1 & 24.5 & 74.0 & 32.7 & 57.9 & 21.3 & 64.7 & 19.6 & 47.5 \\
    UI-TARS-1.5-7B~\citep{ui-tars-15-seed} & - & - & - & - & - & - & - & - & - & - & - & - & - & - & 49.6 \\
    GTA1-7B$\dagger$~\citep{yang2025gta1guitesttimescaling} & 53.3 & {17.2} & 66.9 & { \ul 20.7} & { \ul 62.6} & 18.2  & {76.4} & {31.8} & {82.5} & {\ul 50.9} & {48.6} & 25.9 & {65.5} & 25.2 & 50.1\\
    GTA1-72B~\citep{yang2025gta1guitesttimescaling} & 56.9 & {28.1} & { \ul 79.9} & \textbf{33.1} & \textbf{73.2} & {\ul 20.3} & {\ul 81.9} & {\ul 38.2} & \textbf{85.3} & { 49.1} & {\ul 73.8} & \textbf{39.1} & {\ul 74.5} & {\ul 32.5} & 58.4 \\
    \midrule
    \rowcolor{gray!15}
    \multicolumn{16}{l}{\textit{Ours}} \\
    UI-Venus-Ground-7B & {\ul 60.4} & 21.9  & {74.7} & 24.1  & 63.1  & {14.7} & {76.4} & 31.8  & 75.7  & 41.5  & {49.5} & 22.5  & {67.1} & 24.3  & 50.8 \\
    UI-Venus-Ground-72B & \textbf{66.5}  & {\ul 29.7}  & \textbf{84.4}  & \textbf{33.1} & \textbf{73.2} & \textbf{30.8}  & \textbf{84.7}  & \textbf{42.7} & {\ul 83.1} & \textbf{60.4}  & \textbf{75.7}  & {\ul 36.0} & \textbf{77.4}  & \textbf{36.8} & \textbf{61.9} \\
    \bottomrule
    \end{tabular*} 
    \caption{Performance comparison of different agent models across various task categories based on Text, Icon, and Average scores on \textbf{ScreenSpot-Pro}. Note that models with * are reproduced and the $\dagger$ means trained from UI-TARS-1.5-7B.}
    \label{tab:screenspot_pro}
\end{table*}
\begin{table*}[t]
    \centering
    \footnotesize
    \setlength{\tabcolsep}{0pt}
    \begin{tabular*}{\textwidth}{@{\extracolsep{\fill}}l *{6}{c}}
    \toprule
    \textbf{Models} & 
    \textbf{\begin{tabular}[c]{@{}c@{}}Text \\ Matching\end{tabular}} & 
    \textbf{\begin{tabular}[c]{@{}c@{}}Element \\ Recognition\end{tabular}} & 
    \textbf{\begin{tabular}[c]{@{}c@{}}Layout \\ Understanding\end{tabular}}& 
    \textbf{\begin{tabular}[c]{@{}c@{}}Fine-grained \\ Manipulation\end{tabular}}& 
    \textbf{Refusal} & 
    \textbf{Avg} \\
    \midrule
    \rowcolor{gray!15}
    \multicolumn{7}{l}{\textit{Closed-source Models}} \\
    Operator~\citep{openai2025operator}                        &  51.3 & 42.4&  46.6&  31.5&  - & 40.6      \\ 
    Gemini-2.5-Pro~\citep{geminiteam2024gemini15unlockingmultimodal}                 &59.8 &45.5& 49.0 &33.6& \textbf{38.9} &45.2      \\       
    Seed1.5-VL~\citep{seed2025seed1_5vl}                      &73.9 &66.7 &69.6 &47.0 &{ \ul 18.5} &62.9     \\    
    \midrule
    \rowcolor{gray!15}
    \multicolumn{7}{l}{\textit{Open-source Models}} \\
    OS-Atlas-7B~\citep{wu2024osatlasfoundationactionmodel}   & 44.1 &29.4 &35.2& 16.8 &7.4 &27.7            \\
    Qwen2.5-VL-7B~\citep{bai2025qwen25vltechnicalreport}    &45.6 &32.7 &41.9& 18.1& - & 31.4            \\ 
    Qwen2.5-VL-72B~\citep{bai2025qwen25vltechnicalreport} &52.6 &74.6 &{ \ul 74.7} &55.3 & - &62.2       \\
    UGround-7B~\citep{gou2024navigatingdigitalworldhumans}     &51.3& 40.3& 43.5& 24.8 &- &36.4       \\ 
    Aguvis-7B~\citep{xu2024aguvis}   &  55.9 &41.2& 43.9& 28.2& - & 38.7   \\
    Jedi-7B~\citep{xie2025scalingcomputerusegroundinguser}       & 65.9 &55.5& 57.7& 46.9& 7.4& 54.1        \\ 
    UI-TARS-7B~\citep{qin2025uitarspioneeringautomatedgui} &60.2& 51.8& 54.9 &35.6 & - &47.5       \\
    UI-TARS-1.5-7B~\citep{ui-tars-15-seed}   & 52.6 &75.4 &72.4 & {\ul 66.7} & - &64.2        \\
    UI-TARS-72B~\citep{qin2025uitarspioneeringautomatedgui} & 69.4 &60.6 &62.9 &45.6& - &57.1    \\
    GTA1-7B~\citep{yang2025gta1guitesttimescaling}   &63.2 &\textbf{82.1} &74.2 & \textbf{70.5} & - & { \ul 67.7}           \\
    GTA1-72B~\citep{yang2025gta1guitesttimescaling}  &57.9 &{\ul 76.9} &\textbf{77.3} & { \ul 66.7} & - &66.7    \\
    OpenCUA-7B~\citep{wang2025opencuaopenfoundationscomputeruse}  &- &- &- & - & - &55.3    \\ 
    OpenCUA-32B~\citep{wang2025opencuaopenfoundationscomputeruse}  &-&- &- & - & - &59.6    \\
    \midrule
    \rowcolor{gray!15}
    \multicolumn{7}{l}{\textit{Ours}} \\
    UI-Venus-Ground-7B                       &{\ul 74.6}  &60.5   &61.5  &45.5  & - & 58.8    \\ 
    UI-Venus-Ground-72B                      &\textbf{82.1}  &71.2   &70.7  &64.4  & - & \textbf{70.4}    \\ 
    \bottomrule
    \end{tabular*}
    \vspace{0.4em}
    \caption{Performance comparison on \textbf{OS-World-G} grounding dataset. Our UI-Venus-72B achieves state-of-the-art performance with a significant lead on the Text Matching subtask.}
    \label{tab:osworld_g}
\end{table*}

\begin{table}[ht]
    \centering
    \footnotesize
    \begin{tabular}{lcccc}
    \toprule
    \textbf{Models} & \textbf{Basic} & \textbf{Functional} & \textbf{Spatial}   & \textbf{Avg}                  \\ 
    \midrule
    \rowcolor{gray!15}
    \multicolumn{5}{l}{\textit{Closed-source Models}} \\
    GPT-4o~\citep{openai2024gpt4o}                           & 1.6          & 1.5                 &            1.0       &  1.4    \\ 
    Claude-3.7-Sonnet~\citep{anthropic2024cuda}  & 9.5          & 7.7                &            7.6       &  8.3 \\
    \midrule
    \rowcolor{gray!15}
    \multicolumn{5}{l}{\textit{Open-source Models}} \\
    Qwen2.5-VL-7B~\citep{bai2025qwen25vltechnicalreport}         &    1.2         &     0.8    &       0.5      &      0.9               \\ 
    SeeClick~\citep{cheng2024seeclickharnessingguigrounding}      &   9.4 & 4.7&  2.1&  5.4     \\ 
    UGround-V1-7B~\citep{gou2024navigatingdigitalworldhumans}       &      15.4  & 17.1  & 6.3 &  12.9           \\ 
    OS-Atlas-7B~\citep{wu2024osatlasfoundationactionmodel}    &12.2 &11.2 &3.7 &9.0            \\
    UI-TARS-7B~\citep{qin2025uitarspioneeringautomatedgui}       & 20.1 &24.3 &8.4 &17.6                \\
    UI-TARS-1.5-7B~\citep{ui-tars-15-seed}      & 28.8 &27.5& 10.7 &22.3              \\
    UI-TARS-72B~\citep{qin2025uitarspioneeringautomatedgui}       & 31.4 &30.5 & {\ul 14.7} & 25.5                \\
    Phi-Ground-7B-16C-DPO~\citep{phiground}&  { \ul 36.8} & { \ul 37.1} & 7.6 & { \ul 27.2}              \\
    \midrule
    \rowcolor{gray!15}
    \multicolumn{5}{l}{\textit{Ours}} \\
    UI-Venus-Ground-7B                        &36.1  &32.8        &11.9         &    26.5  \\ 
    UI-Venus-Ground-72B                        & \textbf{45.6} & \textbf{42.3}        & \textbf{23.7}        &    \textbf{36.8}   \\ 
    \bottomrule
    \end{tabular}
    \vspace{0.4em}
    \caption{Performance comparison on \textbf{UI-Vision} grounding dataset. Our UI-Venus-Ground-72B achieves state-of-the-art performance, outperforming all baseline methods across different settings.}
    \label{tab:ui-vison}
\end{table}

\begin{table}[ht]
    \centering
    \footnotesize
    \begin{tabular}{lccc}
    \toprule
    \textbf{Models} & \textbf{Fun2Point} & \textbf{Text2Point}    & \textbf{Avg}                  \\ 
    \midrule
    \rowcolor{gray!15}
    \multicolumn{4}{l}{\textit{Closed-source Models}} \\
    GPT-4o~\citep{openai2024gpt4o}                           & 22.1          & 19.9                 &            21.0             \\ 
    GPT-4o w grounding~\citep{openai2024gpt4o}                           & 44.3          & 44.0                 &   44.2    \\        
    \midrule
    \rowcolor{gray!15}
    \multicolumn{4}{l}{\textit{Open-source Models}} \\
    Intern2.5-VL-8B~\citep{chen2024expanding}                      & 17.2          & 24.2                          &     20.7        \\ 
    Intern2.5-VL-26B~\citep{chen2024expanding}                      & 14.8          & 16.6                        &        15.7      \\ 
    Qwen2.5-VL-7B~\citep{bai2025qwen25vltechnicalreport}                      & 59.8          & 59.3                &   59.6                     \\ 
    OS-Genesis-7B~\citep{osgenesis}                        & 8.3          & 5.8                 &   7.1    \\
    OS-Atlas-7B~\citep{wu2024osatlasfoundationactionmodel}                 & 53.6          & 60.7                 &   57.2           \\
    Aguvis-7B~\citep{xu2024aguvis}                        & 60.8          & 76.5                 &                68.7         \\
    UI-TARS-7B~\citep{qin2025uitarspioneeringautomatedgui}                        & 56.8          & 66.7                 &         61.8                 \\
    AgentCPM-GUI~\citep{agentcpm}                        & 79.1&76.5                &        77.8                \\
    \midrule
    \rowcolor{gray!15}
    \multicolumn{4}{l}{\textit{Ours}} \\
    UI-Venus-Ground-7B                        & { \ul 83.3} &  { \ul 83.2} & { \ul 83.3}         \\ 
    UI-Venus-Ground-72B                        & \textbf{84.1} & \textbf{85.9}        & \textbf{85.0}               \\ 
    \bottomrule
    \end{tabular}
    \vspace{0.4em}
    \caption{Performance comparison on \textbf{CA-GUI} grounding dataset. Our UI-Venus-72B achieves state-of-the-art performance, outperforming all baseline methods across different settings.}
    \label{tab:AgentCPM-GUI}
\end{table}

We evaluate UI-Venus on five comprehensive GUI grounding benchmarks to assess its ability to associate natural language instructions with corresponding GUI elements, including ScreenSpot-V2~\cite{wu2024osatlasfoundationactionmodel}, ScreenSpot-Pro~\cite{li2024screenspot-pro}, OSWorld-G~\cite{xie2025scalingcomputerusegroundinguser}, UI-Vision~\cite{nayak2025ui} and CA-GUI~\cite{agentcpm}. 
During the evaluation, we follow the standard protocol~\cite{cheng2024seeclickharnessingguigrounding,lin2024showuivisionlanguageactionmodelgui}, \emph{i.e.}, a prediction is considered correct when the center of predicted box falls within the ground truth bounding box. 

In the experiments, we compare UI-Venus models against various state-of-the-art baselines across different model categories:
\textbf{(1) Closed-source Models}: UI-TARS-1.5~\cite{ui-tars-15-seed}, Seed1.5-VL~\cite{seed2025seed1_5vl}, GPT-4o~\cite{openai2024gpt4o}, and Claude Computer Use~\cite{anthropic2024cuda}.
\textbf{(2) General Open-source Models}: Qwen2.5-VL-7B/72B~\cite{bai2025qwen25vltechnicalreport} and InternVL2.5~\cite{chen2024expanding}.
\textbf{(3) GUI-specific SFT Models}: UI-TARS-7B/72B~\cite{qin2025uitarspioneeringautomatedgui}, OS-Atlas-7B~\cite{wu2024osatlasfoundationactionmodel}, Jedi-7B~\cite{xie2025scalingcomputerusegroundinguser}, and GUI-Actor-7B~\cite{guiactor}.
\textbf{(4) GUI RL Models}: GTA1-7B/72B~\cite{yang2025gta1guitesttimescaling}, SE-GUI-7B~\cite{yuan2025enhancingvisualgroundinggui}, and UI-TARS-1.5-7B~\cite{ui-tars-15-seed}.
Methods marked with * indicate our own reproductions, some of which achieve superior performance compared to their original reported results.

\noindent \textbf{ScreenSpot-V2.} As a general GUI grounding benchmark across mobile, web and desktop platforms with text and icon/widget elements, this benchmark tests the basic grounding capabilities in daily GUI-related scenarios. From Table~\ref{tab:screenspot-v2}, it can be observed that though previous models have already obtained high scores, UI-Venus-Ground-72B further pushes the performance boundary with 95.3\% average score and exhibits superior grounding ability in diverse scenarios of daily life. Notably, UI-Venus-Ground-7B, with significantly fewer parameters, can still achieve 94.1\% and outperform larger models like UI-TARS-72B or OpenCUA-32B, which only score 90.3\% and 93.4\%, respectively.

\noindent \textbf{ScreenSpot-Pro.} With high-resolution professional software interfaces, including CAD, development, creative, scientific, office, and OS applications, the difficulty of element grounding is greatly increased. Nevertheless, as shown in Table~\ref{tab:screenspot_pro}, UI-Venus-Ground-72B can achieve comprehensive leadership with 61.9\% score. Compared to text elements, we can find in professional software, icon elements are much harder to detect and locate due to their diverse and small shapes, which requires the model's understanding of GUI layouts and fine-grained coordinate perception, while UI-Venus-Ground-72B still makes a breakthrough in icon grounding for software of `Creative', `Scientific' and `Office' categories, with improvements over previous optimal model GTA1-72B of 10.5\%, 4.5\% and 9.5\%, respectively.

\noindent \textbf{OSWorld-G.} This benchmark is composed of fine-grained tasks from real computer environment with diverse capabilities including text matching, element recognition, layout understanding, fine-grained manipulation and refusal handling. The result in Table~\ref{tab:osworld_g} demonstrates our model's exceptional performance on the text matching subtask, where 72B model achieves 70.4\% average score, substantially surpassing strong baselines like UI-TARS-1.5-7B and GTA1 series. 

\noindent \textbf{UI-Vision.} UI-Vision provides a fine-grained evaluation of computer-using agents in real-world desktop environments, serving as a comprehensive, license-permissive benchmark. In Table~\ref{tab:ui-vison}, we can observe UI-Venus-Ground-72B establishes new state-of-the-art performance across three task categories, while UI-Venus-Ground-7B performs on par with previous best model Phi-Ground-7B-16C-DPO, indicating our model has made significant progress in grounding capability under real-world environments.

\noindent \textbf{CA-GUI.} We also adopt multilingual evaluation with realistic Chinese mobile applications, featuring Fun2Point and Text2Point tasks, and the results are exhibited in Table~\ref{tab:AgentCPM-GUI}. In this benchmark, our model demonstrates strong cross-lingual generalization capabilities in non-English GUI environments, markedly exceeding AgentCPM-GUI by 5\% in Fun2Point task and by 9.4\% in Text2Point tasks, and even UI-Venus-Ground-7B shows great advantages.

In summary, our evaluation reveals several key insights: \textbf{(1) Consistent SOTA Performance:} UI-Venus achieves new state-of-the-art results across all benchmarks, with UI-Venus-Ground-72B reaching 95.3\% on ScreenSpot-V2, 61.9\% on ScreenSpot-Pro, and 85.0\% on CA-GUI, significantly outperforming previous best models including GTA1-72B (94.8\%, 58.4\%, -) and UI-TARS-1.5 (95.2\%, 61.6\%, -). \textbf{(2) Superior Fine-grained Capabilities:} On OSWorld-G, our model demonstrates exceptional performance in the text matching tasks, substantially surpassing strong baselines like UI-TARS-1.5-7B and GTA1 series on average. \textbf{(3) Robust Multilingual Grounding:} UI-Venus shows substantial improvements on the Chinese benchmark CA-GUI, achieving 83.2\% average accuracy and demonstrating strong cross-lingual generalization capabilities in non-English GUI environments.

\begin{table}[ht]
    \centering
    \footnotesize
    \begin{tabular}{lcccc}
    \toprule
    \textbf{Models} &\textbf{With Planner} &\textbf{A11y Tree}  &\textbf{Screenshot}  & \textbf{Success Rate(pass@1)}                  \\ 
    \midrule
    \rowcolor{gray!15}
    \multicolumn{5}{l}{\textit{Closed-source Models}} \\
    GPT-4o~\citep{openai2024gpt4o}                      &  \XSolidBold    & \CheckmarkBold &  \XSolidBold         & 30.6                             \\ 
    ScaleTrack~\citep{huang2025scaletrack}                      &  \XSolidBold    & \CheckmarkBold &  \XSolidBold         & 44.0                            \\ 
    SeedVL-1.5~\citep{seed2025seed1_5vl}                      &  \XSolidBold    & \CheckmarkBold &  \CheckmarkBold         & 62.1                         \\ 
    UI-TARS-1.5~\citep{ui-tars-15-seed}                      &  \XSolidBold    & \XSolidBold &  \CheckmarkBold         & 64.2                         \\ 
    \midrule
    \rowcolor{gray!15}
    \multicolumn{5}{l}{\textit{Open-source Models}} \\
    GUI-Critic-R1-7B~\citep{wanyan2025look}                      &   \XSolidBold & \CheckmarkBold & \CheckmarkBold   &  27.6                         \\ 
    Qwen2.5-VL-72B~\citep{bai2025qwen25vltechnicalreport}                      &   \XSolidBold  &  \XSolidBold  & \CheckmarkBold         & 35.0                      \\
    UGround~\citep{gou2024navigatingdigitalworldhumans}                        & \CheckmarkBold     &\XSolidBold & \CheckmarkBold  & 44.0             \\
    Aria-UI~\citep{yang2024ariauivisualgroundinggui} &                \CheckmarkBold     &\XSolidBold & \CheckmarkBold  & 44.8         \\
    UI-TARS-72B~\citep{qin2025uitarspioneeringautomatedgui}               &   \XSolidBold  &  \XSolidBold  & \CheckmarkBold         & 46.6  \\
    GLM-4.5v~\citep{glm-4.5v}       &   \XSolidBold  &   \XSolidBold & \CheckmarkBold         & 57.0  \\
    \midrule
    \rowcolor{gray!15}
    \multicolumn{5}{l}{\textit{Ours}} \\
    UI-Venus-Navi-7B              &  \XSolidBold  &  \XSolidBold    & \CheckmarkBold & 49.1             \\ 
    UI-Venus-Navi-72B              &  \XSolidBold  &   \XSolidBold   & \CheckmarkBold      & \textbf{65.9}               \\ 
    \bottomrule
    \end{tabular}
    \vspace{0.4em}
    \caption{Performance comparison on \textbf{AndroidWorld} for end-to-end models. Our UI-Venus-Navi-72B achieves state-of-the-art performance, outperforming all baseline methods across different settings.}
    \label{tab:Androidworld}
\end{table}

\begin{table}[t]
    \centering
    \footnotesize
    \begin{tabular}{l *{6}{c}}
    \toprule
    \multirow{2}{*}{\textbf{Models}} & \multicolumn{2}{c}{\textbf{AndroidControl-Low}}  & \multicolumn{2}{c}{\textbf{AndroidControl-High}} & \multicolumn{2}{c}{\textbf{GUI-Odyssey}}  \\ 
    \cmidrule(lr){2-3} \cmidrule(lr){4-5} \cmidrule(lr){6-7}
                                     & \textbf{Type Acc.} & \textbf{Step SR} & \textbf{Type Acc.}  & \textbf{Step SR} & \textbf{Type Acc.}  & \textbf{Step SR}                          \\ 
    \midrule
    \rowcolor{gray!15}
    \multicolumn{7}{l}{\textit{Closed-source Models}} \\
    GPT-4o~\citep{openai2024gpt4o}    & 74.3  &19.4 &66.3  &20.8& 34.3& 3.3                              \\ 
    \midrule
    \rowcolor{gray!15}
    \multicolumn{7}{l}{\textit{Open-source Models}} \\
    Qwen2.5-VL-7B~\citep{bai2025qwen25vltechnicalreport}       & 94.1 &85.0 &75.1& 62.9 &59.5& 46.3                           \\ 
    SeeClick~\citep{cheng2024seeclickharnessingguigrounding}     &    93.0 & 75.0&  82.9&    59.1&  71.0 & 53.9                         \\
    OS-Atlas-7B~\citep{wu2024osatlasfoundationactionmodel}   & 93.6 &85.2& 85.2 &71.2& 84.5&62.0                               \\
    Aguvis-7B~\citep{xu2024aguvis}                       & - & 80.5 & - & 61.5  & -   &  -                \\
    Aguvis-72B~\citep{xu2024aguvis}                       & - & 84.4 & - & 66.4  &-  &  -                   \\
    OS-Genesis-7B~\citep{xu2024aguvis}                   & 90.7 & 74.2  & 66.2 & 44.5  &    -                  \\
    UI-TARS-7B~\citep{qin2025uitarspioneeringautomatedgui}           & {\ul98.0} & 90.8 & 83.7 &  72.5   & {\ul94.6} & {\ul87.0}                   \\
    UI-TARS-72B~\citep{qin2025uitarspioneeringautomatedgui}          & \textbf{98.1} & 91.3 & 85.2 & 74.7   & \textbf{95.4} & \textbf{88.6}                       \\

    GUI-R1-7B~\citep{luo2025guir1generalistr1style}         & 85.2 &66.5 &71.6 &51.7 &65.5 &38.8                     \\
    NaviMaster-7B~\citep{navimaster}  & 85.6 & 69.9 & 72.9 & 54.0 & - & -                     \\
    UI-AGILE-7B~\citep{uiagile}          & 87.7 & 77.6 & 80.1 & 60.6 & - & -                     \\
    AgentCPM-GUI~\citep{agentcpm}          & 94.4 & 90.2 & 77.7 & 69.2 & 90.0 & 75.0                     \\
    \midrule
    \rowcolor{gray!15}
    \multicolumn{7}{l}{\textit{Ours}} \\
    UI-Venus-Navi-7B                       &97.1 & {\ul92.4} & \textbf{86.5} & {\ul76.1} & 87.3 &71.5          \\ 
    UI-Venus-Navi-72B                       &96.7  &\textbf{92.9} & {\ul85.9}  & \textbf{77.2} & 87.2 & 72.4             \\ 
    \bottomrule
    \end{tabular}
    \vspace{0.4em}
    \caption{Performance comparison on offline UI navigation datasets including \textbf{AndroidControl} and \textbf{GUI-Odyssey}.}
    \label{tab:navi_off_exp}
\end{table}

\subsection{Navigation Benchmarks}


 We evaluate UI-Venus on three widely-adopted benchmarks: AndroidWorld~\cite{androidworld}, AndroidControl~\cite{androidcontrol}, and GUI-Odyssey~\cite{guiodyssey} datasets to assess its multi-step decision-making capabilities across diverse mobile interface scenarios. Methods marked with * indicate our own reproductions, some of which achieve superior performance compared to their original reported results.

\vspace{0.3em}
\noindent \textbf{Online Benchmark.} For real-time multi-step decision-making evaluation, we adopt AndroidWorld, a dynamic benchmark requiring continuous interaction with live mobile applications. This framework assesses the model's ability to adapt strategies based on real-time feedback and maintain task coherence across extended interaction sequences. 
From Table~\ref{tab:Androidworld}, we can see UI-Venus is implemented without relying on additional planner or A11y tree and is able to conduct end-to-end UI navigation purely based on screenshots, which ensures strong generalization capability in real-time interactive scenarios. In AndroidWorld benchmark, UI-Venus-Navi-72B gains a score of 65.9\% success rate in one-time evaluation, surpassing UI-TARS-1.5 (64.2\%), and UI-Venus-Navi-7B also succeeds with 49.1\% rate to outperform UI-TARS-72B (46.6\%), demonstrating the effectiveness of UI-Venus.

\vspace{0.3em}
\noindent \textbf{Offline Benchmark.} Besides, we employ two established offline benchmarks: AndroidControl and GUI-Odyssey. These static evaluation scenarios assess the model's fundamental UI understanding, task decomposition, and action planning capabilities in controlled environments. With evaluation results in Table~\ref{tab:navi_off_exp}, where `Low/High' categories in AndroidControl represent `the low/high level instruction' the agent is prompted with, respectively, we can observe that UI-Venus produces comparable results with previous optimal methods. Notably, in AndroidControl-High evaluation, UI-Venus performs best in both type accuracy and step success rate, indicating that UI-Venus possesses stronger planning and summary capability with high-level task instruction.

In summary, our evaluation reveals several key insights: \textbf{(1) SOTA Online Navigation Performance:} UI-Venus achieves new state-of-the-art end-to-end model results on AndroidWorld, with UI-Venus-Navi-72B reaching \textbf{65.9\%} success rate, significantly outperforming previous best models including UI-TARS-72B (46.6\%) and UI-TARS-1.5 (64.2\%). We also show some case studies in Sec.~\ref{sec:AndroidWorld_visual} and open-source the all evaluation traces of AndroidWorld to better show the effectiveness of our UI-Venus. \textbf{(2) Comparable Offline Navigation Performance:} On AndroidControl and GUI-Odyssey, our UI-Venus also achieves comparable results with previous SOTA methods. Note that UI-Venus obtains best performance on AndroidControl-High, reflecting its remarkable generalization performance of planning and summary in long traces.

\section{Future Work}


Although the GRPO algorithm has proven to be a more effective post-training strategy for UI agents than conventional supervised fine-tuning, certain challenges remain. One notable issue is the hallucination gap between the model’s internal reasoning (\emph{think}) and its final response (\emph{answer}),
which can lead to incorrect or inconsistent behavior in navigation tasks. How to reduce the hallucination of MLLM is still an open and unsolved problem.

Another observation is that even humans may struggle to operate unfamiliar applications without prior exposure. This suggests that large-scale pretraining on trajectories from diverse UI agents can equip models with richer prior knowledge of various applications and to enhance their adaptability.
Thus, future research may explore integrating pretraining with curated UI-agent traces, refining action generation through enhanced reasoning alignment, and incorporating domain-specific priors. Such advancements would help mitigate these issues and enable broader deployment of UI agents across diverse and dynamic computing environments.

\newpage

\bibliographystyle{assets/plainnat}
\bibliography{main}

\newpage

\begin{appendix}

\section{Prompt Templates}

\subsection{Grounding}\label{sec:gd_prompt}


\begin{tcolorbox}[
    title=Grounding Prompt]
Outline the position corresponding to the instruction: \texttt{\color{red} \{problem\}}. The output should be only [x1,y1,x2,y2].
\end{tcolorbox}

\subsection{Navigation}\label{sec:navi_prompt}

\begin{tcolorbox}[
    title=Navigation Prompt,breakable]
**You are a GUI Agent**. 

Your task is to analyze a given user task, review current screenshot and previous actions, and determine the next action to complete the task.

\medskip

\#\#\# User Task 

\texttt{\color{red} \{problem\}} 

\medskip

\#\#\# Previous Actions 

\texttt{\color{red} \{history\}} 

\medskip

\#\#\# Available Actions 

You may execute one of the following functions:

Click(box=(x1, y1)) 

Drag(start=(x1, y1), end=(x2, y2)) 

Scroll(start=(x1, y1), end=(x2, y2), direction=`down/up/right/left') 

Type(content=`') 

Launch(app=`') 

Wait() 

Finished(content=`') 

CallUser(content=`') 

LongPress(box=(x1, y1)) 

PressBack() 

PressHome() 

PressEnter() 

PressRecent() 

\medskip

\#\#\# Instruction 

- Make sure you understand the task goal to avoid wrong actions. 

- Make sure you carefully examine the current screenshot. Sometimes the summarized history might not be reliable, over-claiming some effects. 

- For requests that are questions (or chat messages), remember to use the `CallUser' action to reply to user explicitly before finishing! Then, after you have replied, use the Finished action if the goal is achieved. 

- Consider exploring the screen by using the `scroll' action with different directions to reveal additional content. 

- To copy some text: first select the exact text you want to copy, which usually also brings up the text selection bar, then click the `copy' button in bar. 

- To paste text into a text box, first long press the text box, then usually the text selection bar will appear with a `paste' button in it. 

- You first think about the reasoning process in the mind, then provide the action. The reasoning and action are enclosed in <think></think> and <action></action> tags respectively. After providing action, summarize your action in <conclusion></conclusion> tags.
\end{tcolorbox}

\section{Qualitative Examples}

\subsection{Grounding}

\begin{figure}[h!]
  \centering
  \includegraphics[width=0.9\linewidth]{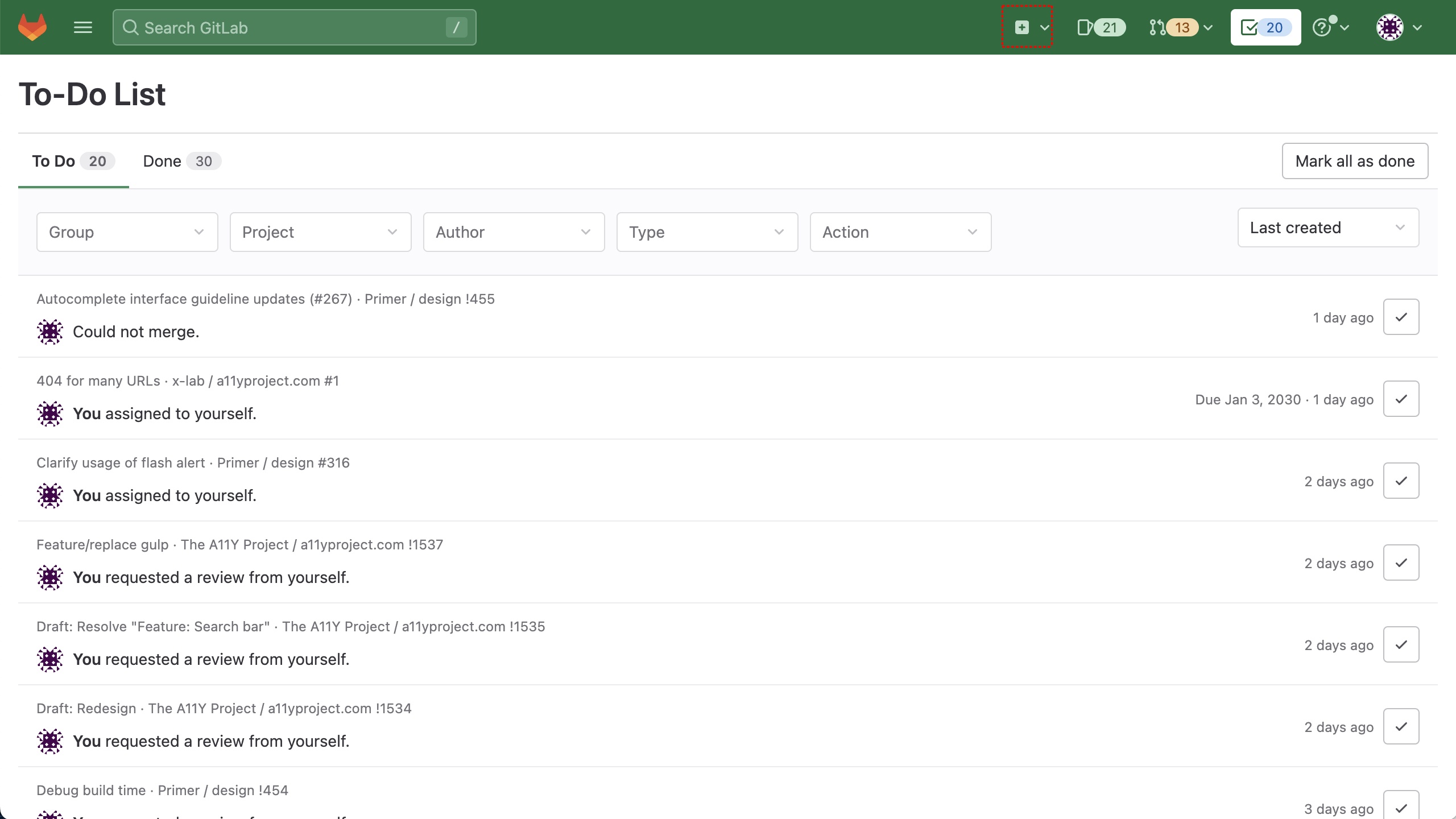}
  \caption{Grounding example for the website Gitlab. The instruction is "Add a new one" and the bounding box result is displayed as a dotted red box.}
  \label{fig:website_gitlab}
\end{figure}

\begin{figure}[h!]
  \centering
  \includegraphics[width=0.9\linewidth]{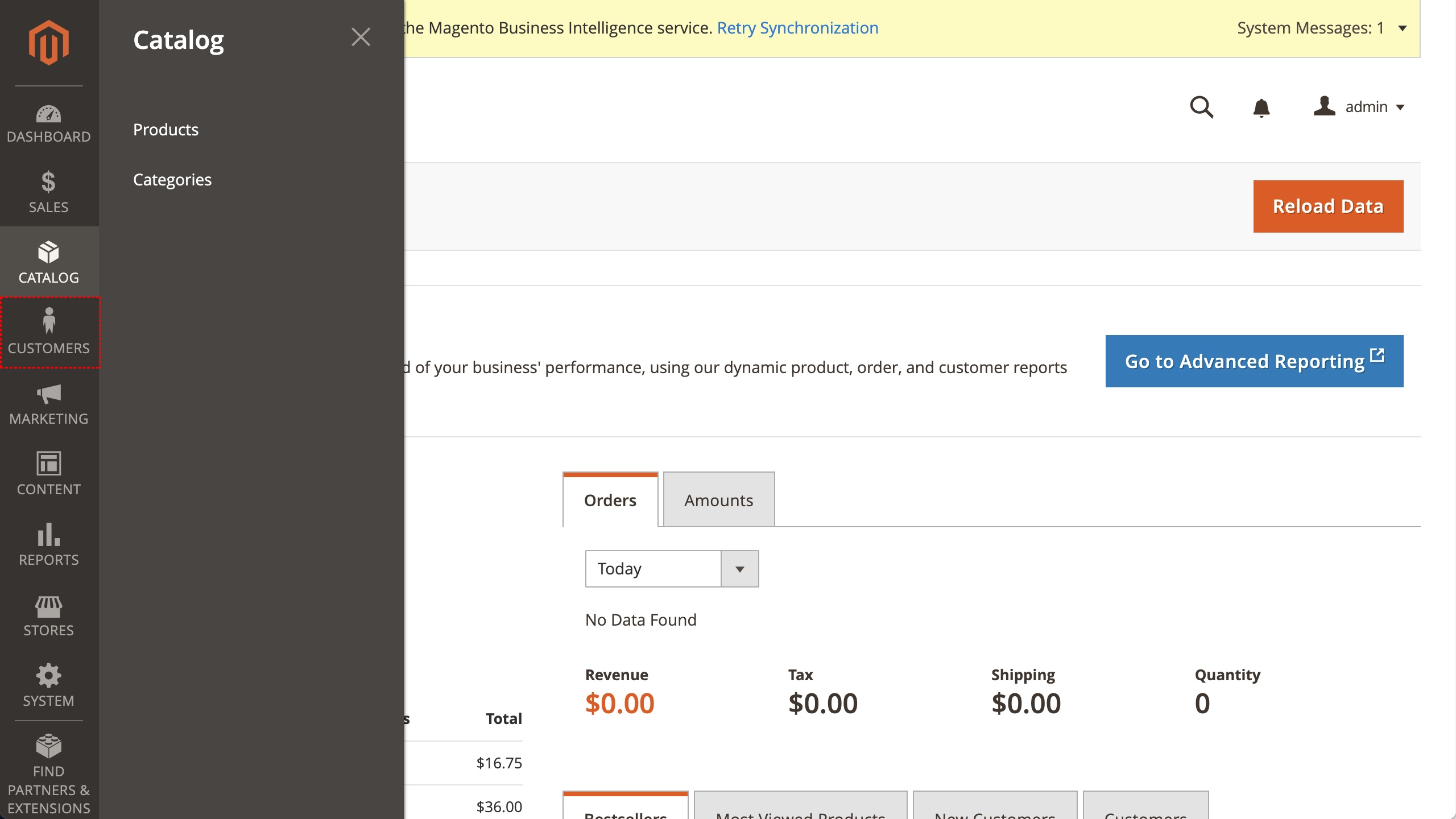}
  \caption{Grounding example for a shopping website. The instruction is "Switch to customers" and the bounding box result is displayed as a dotted red box.}
  \label{fig:website_shop}
\end{figure}

\begin{figure}[h!]
  \centering
  \includegraphics[width=0.95\linewidth]{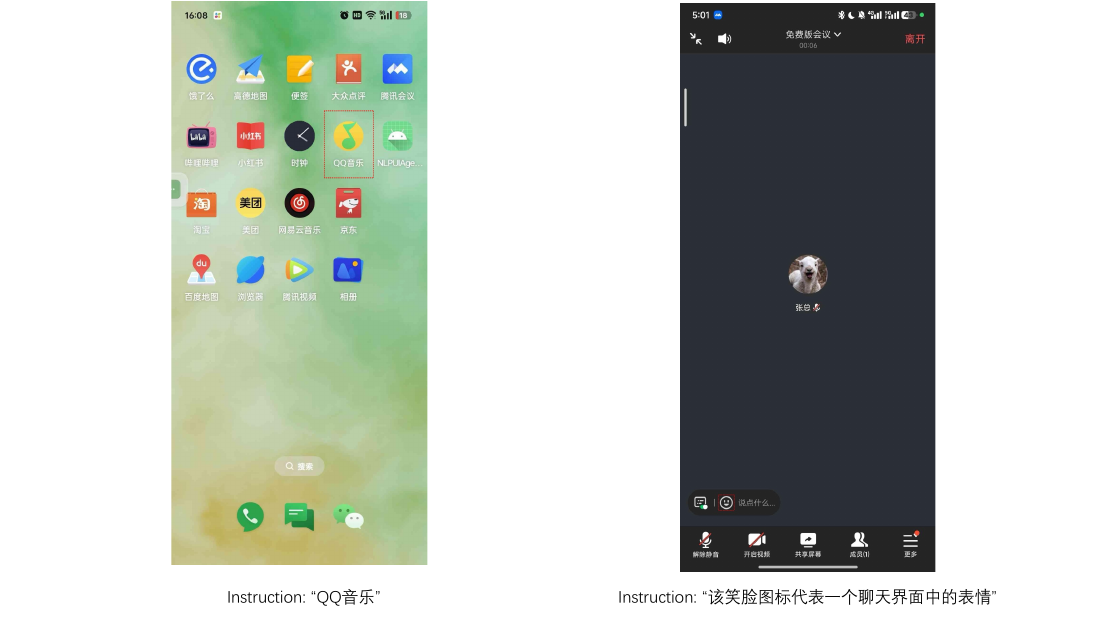}
  \caption{Grounding examples for the mobile system with Chinese instructions. The bounding box results are displayed as dotted red boxes.}
  \label{fig:mobile_music}
\end{figure}

We show several grounding examples across desktop professional software, desktop system, website and mobile in Figure~\ref{fig:architecture_blender}-\ref{fig:mobile_music}.

\begin{figure}[h!]
  \centering
  \includegraphics[width=0.9\linewidth]{grounding_figures/icon_blender_windows_1.jpg}
  \caption{Grounding example for desktop professional software Blender. The instruction is "Increase Z axis" and the bounding box result is displayed as a dotted red box}
  \label{fig:architecture_blender}
\end{figure}

\begin{figure}[h!]
  \centering
  \includegraphics[width=0.9\linewidth]{grounding_figures/icon_excel_macos_1.jpg}
  \caption{Grounding example for Excel. The instruction is "Redo" and the bounding box result is displayed as a dotted red box.}
  \label{fig:architecture_excel}
\end{figure}

\begin{figure}[h!]
  \centering
  \includegraphics[width=0.9\linewidth]{grounding_figures/text_linux_common_linux_1.jpg}
  \caption{Grounding example for common linux system. The instruction is "Change the terminal encoding to the legacy GBK" and the bounding box result is displayed as a dotted red box.}
  \label{fig:architecture_linux}
\end{figure}

\newpage
~\\~\\~\\~\\~\\~\\

\subsection{Navigation}\label{sec:AndroidWorld_visual}

\begin{figure*}[h]
  \centering
  \includegraphics[width=0.9\linewidth]{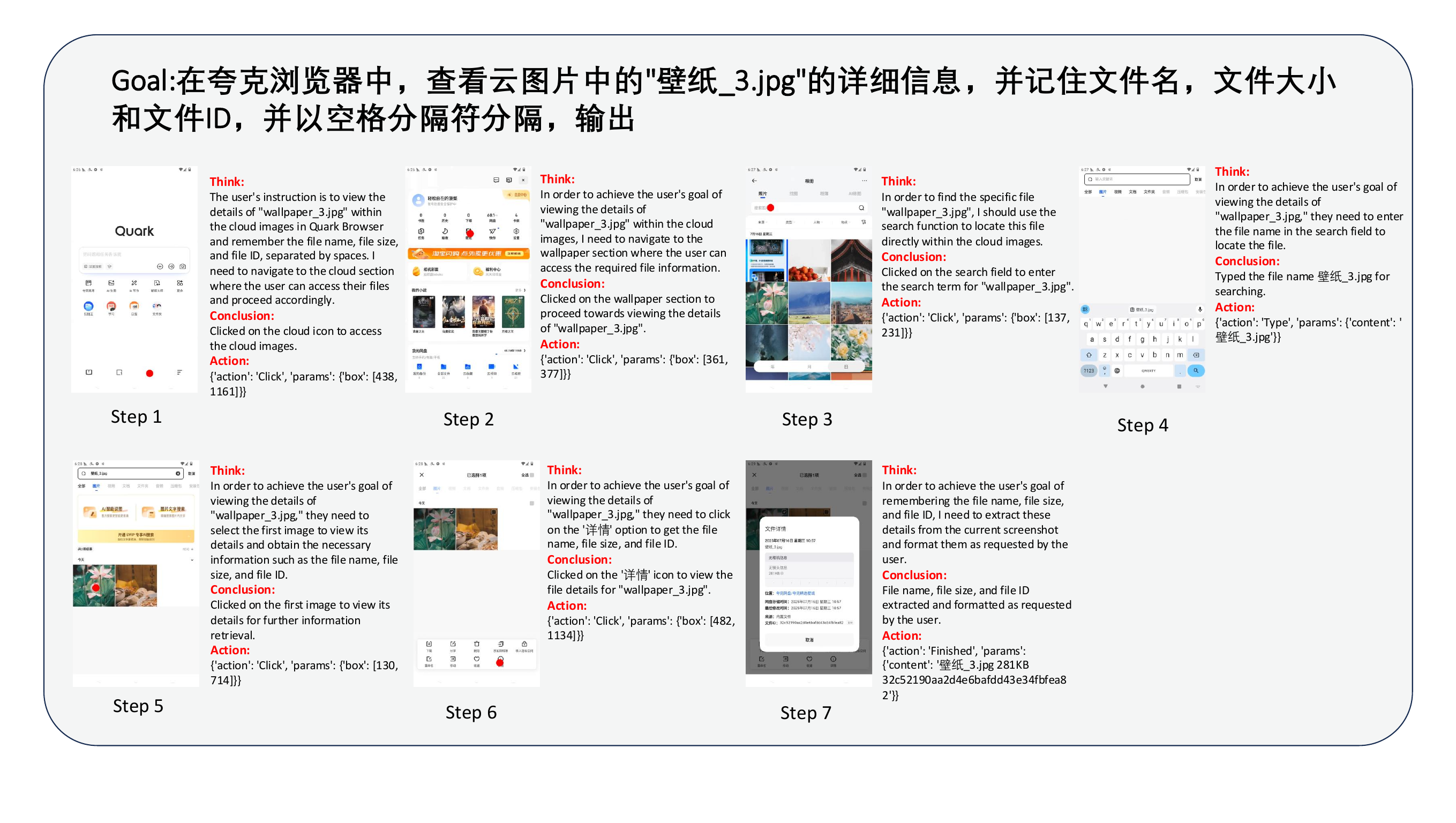}
  \caption{One UI-Venus navigation trace on a Chinese application with Chinese goal description. Interaction points are highlighted using red circular markers.}
  \label{fig:test_trace}
\end{figure*}
\begin{figure*}[h]
  \centering
  \includegraphics[width=0.95\linewidth]{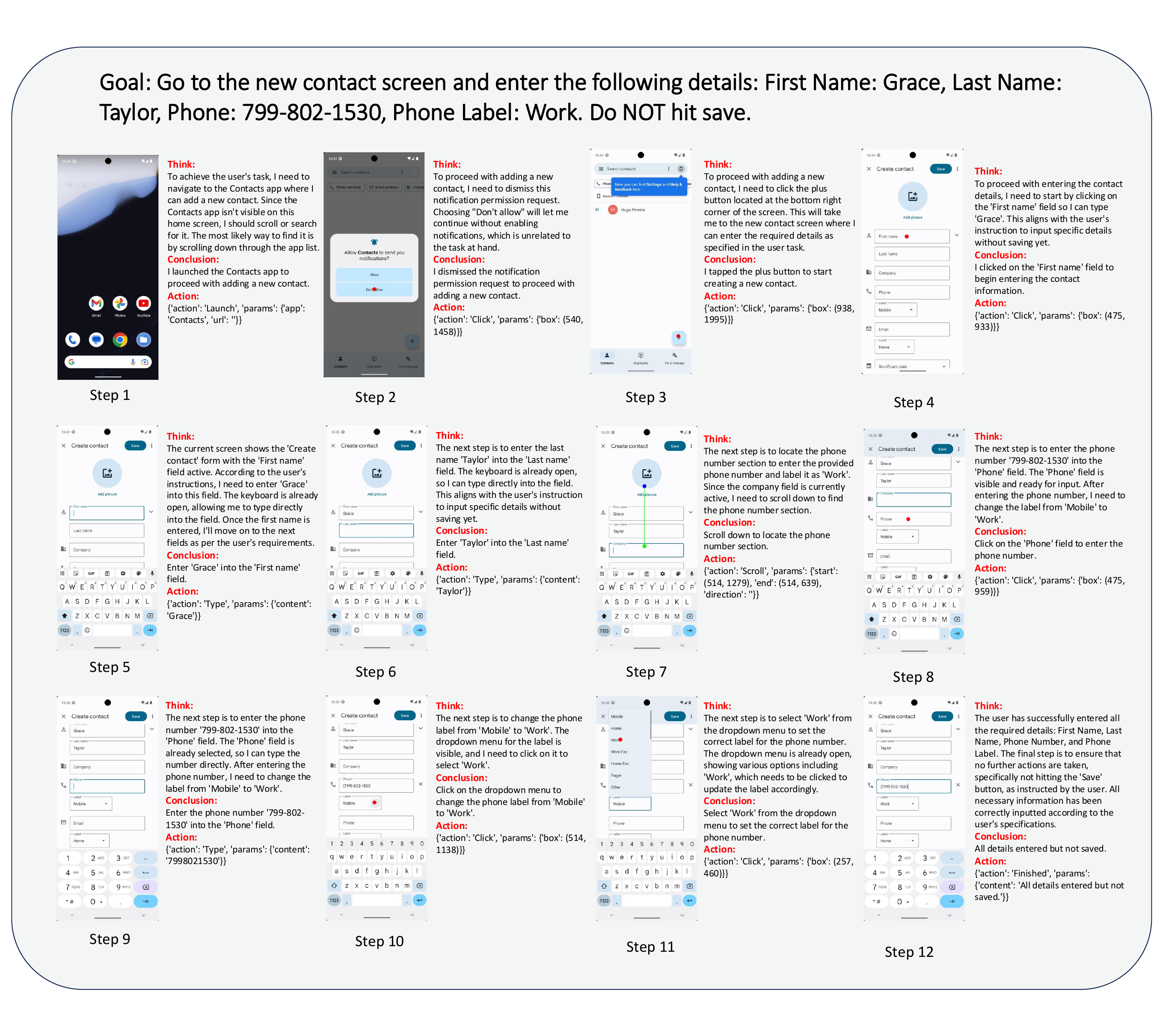}
  \caption{One trace of UI-Venus on the task named \emph{ContactsNewContactDraft} in AndroidWorld. We mark the clicking points with red circles and observe that UI-Venus successfully create the draft without hitting the Save, which shows the model's powerful navigation generalization and strong instruction-following ability.}
  \label{fig:contract}
\end{figure*}
\begin{figure*}[h]
  \centering
  \includegraphics[width=0.95\linewidth]{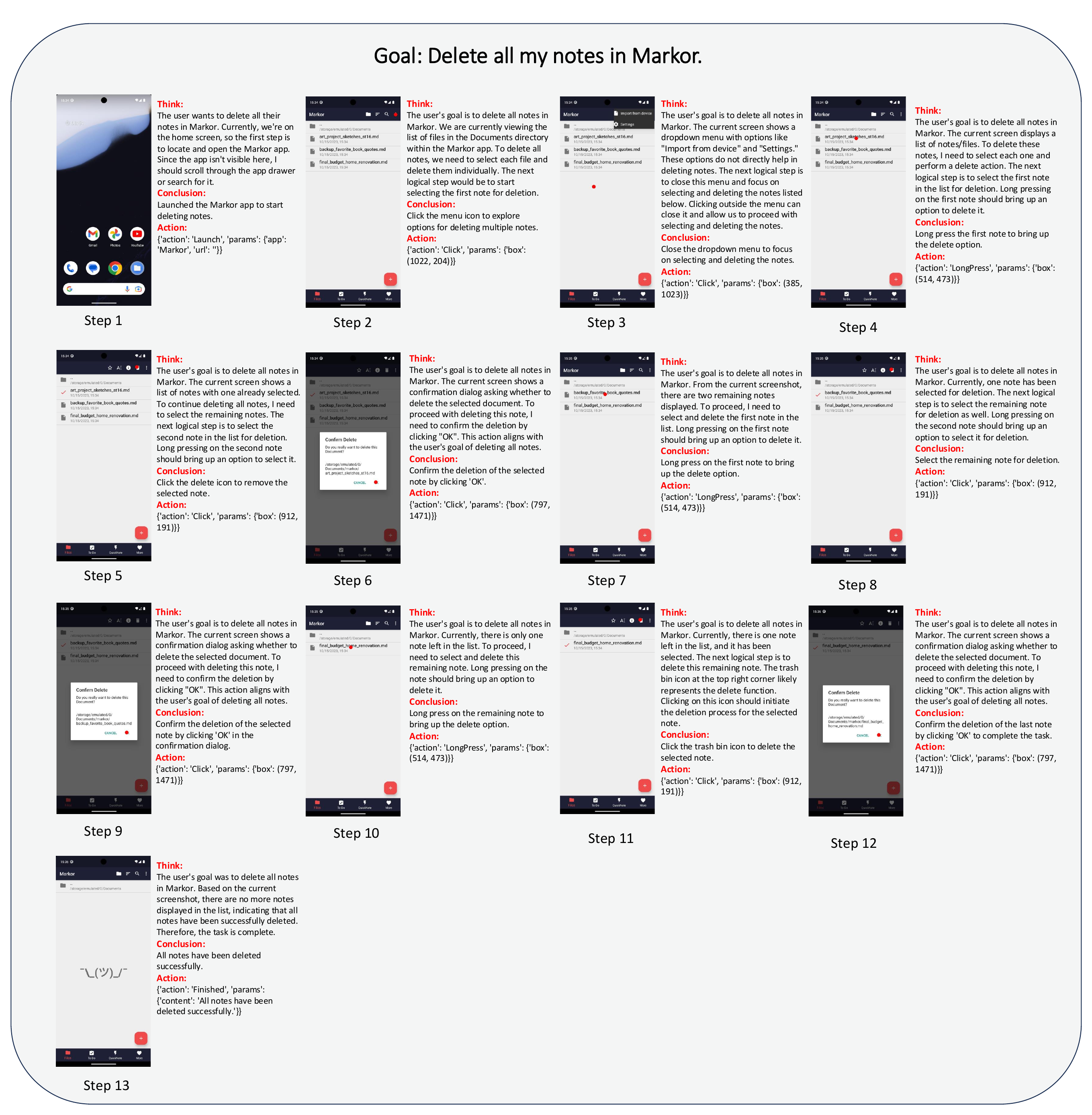}
  \caption{One trace of UI-Venus on the task named \emph{MarkorDeleteAllNotes} in AndroidWorld. We can observe that UI-Venus successfully achieves the goal and has the reflection ability in Step 3. However, there also exists the conflict between think and action in Step 5, remaining as a future work about how to solve MLLM's hallucination.}
  \label{fig:markor}
\end{figure*}

We present several UI navigation case studies in Figures~\ref{fig:contract} and~\ref{fig:markor} from the AndroidWorld dataset. To demonstrate UI-Venus's cross-lingual capabilities, we additionally provide a navigation trace (Figure~\ref{fig:test_trace}) showcasing its performance on a Chinese language application with Chinese interface elements.

\end{appendix}

\end{document}